\documentclass[10pt,twocolumn,letterpaper]{article}

\usepackage{iccv}
\usepackage{times}
\usepackage{epsfig}
\usepackage{graphicx}
\usepackage{amsmath}
\usepackage{amssymb}
\usepackage[dvipsnames]{xcolor}
\usepackage{booktabs}
\usepackage{graphicx}
\usepackage{subcaption}
\usepackage{multirow}
\usepackage{algorithm}
\usepackage{algorithmic}
\usepackage[accsupp]{axessibility} 
\usepackage[sort&compress,numbers]{natbib}

\usepackage{amsmath,amsfonts,bm}

\def\eqref#1{equation~\ref{#1}}

\def\1{\bm{1}}

\def\rvc{{\mathbf{c}}}

\def\rvu{{\mathbf{i}}}

\def\rvp{{\mathbf{p}}}
\def\rvq{{\mathbf{q}}}

\def\rvu{{\mathbf{u}}}
\def\rvv{{\mathbf{v}}}

\def\rvx{{\mathbf{x}}}

\def\rvz{{\mathbf{z}}}

\def\vf{{\bm{f}}}

\def\vx{{\bm{x}}}

\def\mI{{\bm{I}}}

\DeclareMathAlphabet{\mathsfit}{\encodingdefault}{\sfdefault}{m}{sl}
\SetMathAlphabet{\mathsfit}{bold}{\encodingdefault}{\sfdefault}{bx}{n}

\def\gN{{\mathcal{N}}}

\newcommand{\E}{\mathbb{E}}

\newcommand{\KK}[1]{\noindent{\color{magenta}{\textbf{Karsten}: #1}}}

\usepackage[pagebackref=true,breaklinks=true,letterpaper=true,colorlinks,bookmarks=false]{hyperref}

\iccvfinalcopy 

\def\iccvPaperID{11171} 

\ificcvfinal\fi

\newcommand{\ourmodel}{TexFusion }

\begin{document}

\title{\vspace{-2.3em} \textit{TexFusion}: \\ Synthesizing 3D Textures with Text-Guided Image Diffusion Models}

\author{Tianshi Cao $^{1,2,3}$ 
\hspace{0.5cm} Karsten Kreis $^{1}$ 
\hspace{0.5cm} Sanja Fidler$^{1,2,3}$ 
\hspace{0.5cm} Nicholas Sharp$^{1, *}$ 
\hspace{0.5cm} Kangxue Yin$^{1, *}$ 
\\
{\small$^1$ NVIDIA, \quad $^2$ University of Toronto, \quad $^3$ Vector Institute}\\
{\tt\small
  \{tianshic, kkreis, sfidler, nsharp, kangxuey\}@nvidia.com
}
}

\ificcvfinal\thispagestyle{empty}\fi

\twocolumn[{
   \renewcommand\twocolumn[1][]{#1}
   \maketitle
   \begin{center}
       \includegraphics[trim={2.6cm 18cm 13cm 1cm},clip,width=0.78\linewidth]{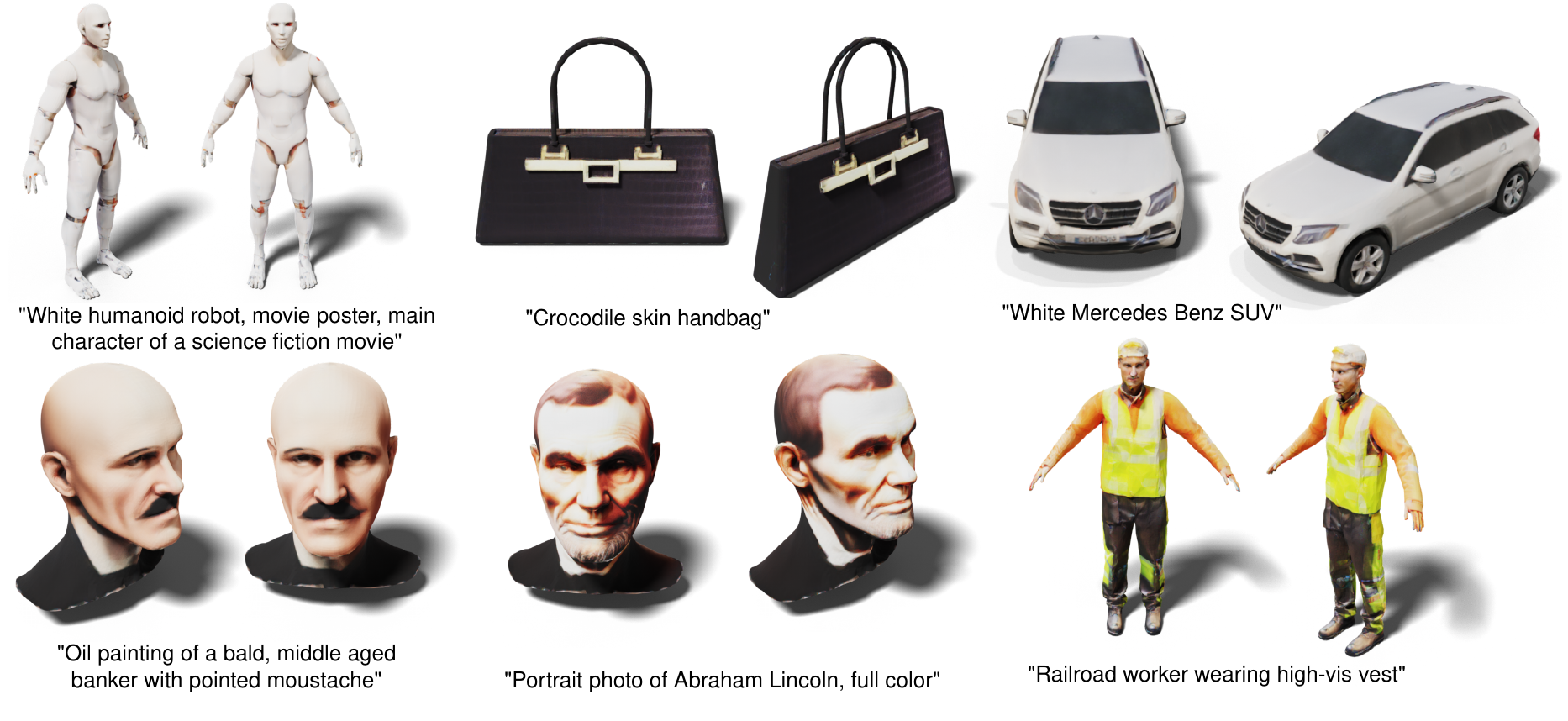}\vspace{-6pt}
\captionof{figure}{\small Text-conditioned 3D texturing results with TexFusion.}
\label{fig:front_figure}

   \end{center}
}]

\let\thefootnote\relax\footnote{$^*$ Equal contribution.}

   

\begin{abstract}
\vspace{-6pt}
  We present \emph{TexFusion} (\emph{Texture Diffusion}), a new method to synthesize textures for given 3D geometries, using large-scale text-guided image diffusion models. In contrast to recent works that leverage 2D text-to-image diffusion models to distill 3D objects using a slow and fragile optimization process, TexFusion introduces a new 3D-consistent generation technique specifically designed for texture synthesis that employs regular diffusion model sampling on different 2D rendered views. Specifically, we leverage latent diffusion models, apply the diffusion model's denoiser on a set of 2D renders of the 3D object, and aggregate the different denoising predictions on a shared latent texture map. Final output RGB textures are produced by optimizing an intermediate neural color field on the decodings of 2D renders of the latent texture. We thoroughly validate TexFusion and show that we can efficiently generate diverse, high quality and globally coherent textures. We achieve state-of-the-art text-guided texture synthesis performance using only image diffusion models, while avoiding the pitfalls of previous distillation-based methods. The text-conditioning offers detailed control and we also do not rely on any ground truth 3D textures for training. This makes our method versatile and applicable to a broad range of geometry and texture types. We hope that TexFusion will advance AI-based texturing of 3D assets for applications in virtual reality, game design, simulation, and more. Videos and more results on \href{https://research.nvidia.com/labs/toronto-ai/texfusion/}{project webpage}.
  \looseness=-1
  \vspace{-0.4cm}
\end{abstract}

\vspace{-0.3cm}
\section{Introduction}
\vspace{-0.2cm}
In the past decade, deep learning-based 3D object generation has been studied extensively~\cite{wu2016learning,achlioptas2018learning,sun2018pointgrow,xie2018learning,mo2019structurenet,valsesia2019learning,li2018point,shu20193d,huang20193d,yang2019pointflow,klokov2020discrete,nash2020polygen,ibing2021shape,li2021spgan,yin2021_3DStyleNet,zhang2021image,xie2021GPointNet,luo2021surfgen,wen2021learning,ko2021rpg,sanghi2021clipforge,gao2022get3d,park2019deepsdf,luo2021diffusion,zhou2021shape,zeng2022lion,nichol2022pointe,kalischek2022tetrahedral,shue2022triplanediffusion,bautista2022gaudi,nam20223dldm,wang2022rodin}, 
due to the demand for high-quality 3D assets in 3D applications such as VR/AR, simulation, digital twins, etc. 
While many prior works on 3D synthesis focus on the geometric components of the assets, textures are studied much less, despite the fact that they are important components of realistic 3D assets which assign colors and materials to meshes to make the rendering vivid. Recent advances in text-conditioned image diffusion models trained on internet-scale data~\cite{rombach2021highresolution,nichol2021glide,saharia2022imagen,ramesh2022dalle2,balaji2022eDiffi} have unlocked the capability to generate images with stunning visual detail and practically unlimited diversity. These high-performance diffusion models have also been used as image priors to synthesize 3D objects with textures using textual guidance~\cite{poole2022dreamfusion,lin2022magic3d,metzer2022latent,wang2022scorejacobian}. 

In this paper, we aim to perform text-driven high-quality 3D texture synthesis for given meshes, by leveraging the information about the appearance of textures 
carried by the image prior of a pre-trained text-to-image diffusion model.
The main workhorse in current text-to-3D methods that leverage 2D image diffusion models is \emph{Score Distillation Sampling (SDS)}~\cite{poole2022dreamfusion}. SDS is used to distill, or optimize, a 3D representation such that its renders are encouraged to be high-likelihood under the image prior. Methods utilizing SDS also share two common limitations in that: 1). a high classifier-free guidance weight~\cite{ho2021classifierfree} is required for the optimization to converge, resulting in high color saturation and low generation diversity; 2). a lengthy optimization process is needed for every sample.\looseness=-1

To address the above issues, we present \textit{Texture Diffusion}, or \textit{TexFusion} for short. \ourmodel is a sampler for sampling surface textures from image diffusion models. Specifically, we use \textit{latent} diffusion models that efficiently autoencode images into a latent space and generate images in that space~\cite{vahdat2021score,rombach2021highresolution}. \ourmodel leverages latent diffusion trajectories in multiple object views, encoded by a shared latent texture map. Renders of the shared latent texture map are provided as input to the denoiser of the latent diffusion model, and the output of every denoising step is projected back to the shared texture map in a 3D-consistent manner. To transform the generated latent textures into RGB textures, we optimize a neural color field on the outputs of the latent diffusion model's decoder applied to different views of the object.
We use the publicly available latent diffusion model Stable-Diffusion with depth conditioning~\cite{rombach2021highresolution} (\texttt{SD2-depth})
as our diffusion backbone.
Compared to methods relying on SDS, \ourmodel produces textures with more natural tone, stronger view consistency, and is significantly faster to sample (3 minutes vs. 30 minutes reported by previous works).\looseness=-1

We qualitatively and quantitatively validate \ourmodel on various texture generation tasks. We find that \ourmodel generates high quality, globally coherent and detailed textures that are well-aligned with the text prompts used for conditioning (e.g. Fig.~\ref{fig:front_figure}). Since we leverage powerful text-to-image diffusion models for texture sampling, we can generate highly diverse textures and are not limited to single categories or restricted by explicit 3D textures as training data, which are limitations of previous works~\cite{oechsle2019texture,siddiqui2022texturify,chen2022AUVNET,gao2022get3d,chan2022efficient}.
In summary, our main contribution is a novel method for 3D texture generation from 2D image diffusion models, that is view-consistent, avoids over-saturation and achieves state-of-the-art text-driven texture synthesis performance.\looseness=-1

\begin{figure*}[t]
\vspace{-0.3cm}
    \centering
    \setlength{\tabcolsep}{0pt}
    {\small
    \begin{tabular}{c c c c c c}
        \includegraphics{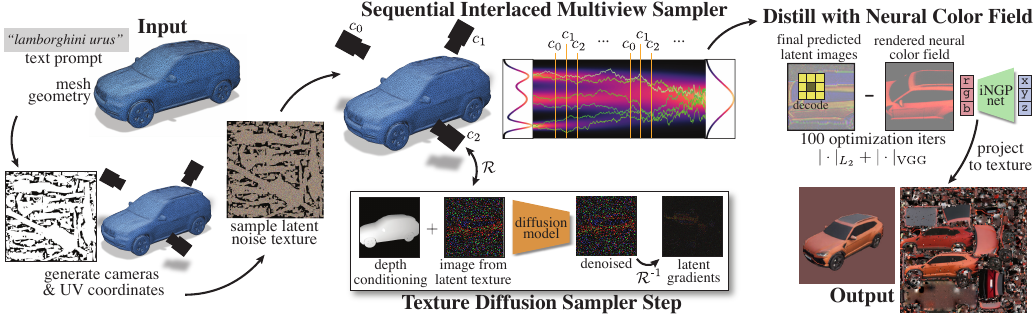}\vspace{-6pt}
    \end{tabular}
    }
    \vspace{-0.2cm}
    \caption{\small Overview of TexFusion. TexFusion takes a text prompt and mesh geometry as input and produces a UV parameterized texture image matching the prompt and mesh. Key to TexFusion is the Sequential Interlaced Multiview Sampler (SIMS) - SIMS performs denoising diffusion iterations in multiple camera views, yet the trajectories are aggregated through a latent texture map after every denoising step. SIMS produces a set of 3D consistent latent images (TexFusion uses Stable Diffusion~\cite{rombach2021highresolution} as text-to-image diffusion backbone), which are decoded and fused into a texture map via optimizing an intermediate neural color field.}
    \label{fig:overview_figure}
    \vspace{-0.4cm}
\end{figure*}

\section{Related Work}

\vspace{-3pt}
\paragraph{Classic Computer Graphics Techniques}
Early work on texture generation focused on tiling exemplar patterns across a surface, often with an explicit direction field for local orientation ~\cite{wei2001texture,turk2001texture,lefebvre2006appearance, kopf2007solid}. 
See \cite{pauly2009survey} for a survey.
This research established the value of texture image representations and the challenges of global coherence, both central to this work.
However, modern learning-based priors have proven necessary to go beyond simple patterns and synthesize plausible shape-specific texture details.

\vspace{-6pt}
\paragraph{Texture Synthesis with 3D Priors} 
Textures are defined on the meshed surface of 3D objects, which is an irregular representation. To enable 2D texture generation on 3D meshes,  
AUV-Net~\cite{chen2022AUVNET} learns an aligned UV space for a set of 3D shapes in a given class, mapping 3D texture synthesis to a 2D domain. 
Texturify~\cite{siddiqui2022texturify} trains a 3D StyleGAN in the quad-parameterized surface domain on a set of textured 3D shapes in a given class.
Different from AUV-NET or Texturify, which embed the geometric prior into a UV map or mesh parameterization,  EG3D~\cite{chan2022efficient} and  GET3D~\cite{gao2022get3d} directly train 3D StyleGANs to generate geometry and texture jointly, where the textures are implicit texture fields~\cite{oechsle2019texture}.
Other works also represent 3D texture by vertex colors~\cite{michel2022text2mesh}, voxels~\cite{chen2019text2shape}, cube mapping~\cite{xiang2021neutex}, etc. In contrast to TexFusion, these works mostly don't offer text conditioning, often work only on single object categories or require textured 3D shapes for training, which limits their broad applicability.

\vspace{-6pt }
\paragraph{Diffusion Models} Our approach directly builds on diffusion models~\cite{yang2022survey,cao2022survey,croitoru2022survey}, which have recently emerged as new state-of-the-art generative models. 
In particular, they have demonstrated outstanding performance in image generation~\cite{dhariwal2021diffusion,ho2021cascaded,dockhorn2022score,dockhorn2022genie,nichol2021improved,vahdat2021score}, outperforming GANs, and led to breakthroughs in text-to-image synthesis~\cite{rombach2021highresolution,nichol2021glide,saharia2022imagen,ramesh2022dalle2,balaji2022eDiffi}. They have also been successfully used for a variety of image editing and processing tasks~\cite{meng2021sdedit,lugmayr2022repaint,saharia2021image,li2022srdiff,sasaki2021unitddpm,saharia2021palette,su2022dual,kawar2022restoration,hertz2022prompt,ruiz2022dreambooth,gal2022animage}.
Moreover 3D object generation has been addressed with diffusion models, too, for instance leveraging point clouds~\cite{luo2021diffusion,zhou2021shape,zeng2022lion,nichol2022pointe}, meshes~\cite{kalischek2022tetrahedral}, or neural fields~\cite{shue2022triplanediffusion,bautista2022gaudi,nam20223dldm,wang2022rodin} as 3D representations. However, these works focus on geometry generation and do not specifically tackle 3D texture synthesis.\looseness=-1

\vspace{-0.3cm}

\paragraph{Distilling 3D Objects from 2D Image Diffusion Models} Recently, large-scale 2D text-to-image diffusion models have been leveraged to \textit{distill} individual 3D objects as neural radiance fields using \textit{Score Distillation Sampling (SDS)}~\cite{poole2022dreamfusion,metzer2022latent,lin2022magic3d,wang2022scorejacobian,deng2022nerdi}. In SDS, the radiance field is rendered from different directions into 2D images and it is optimized such that each render has a high probability under the text-to-image diffusion model while conditioning on a text prompt. DreamFusion~\cite{poole2022dreamfusion} pioneered the approach, Magic3D~\cite{lin2022magic3d} proposed a coarse-to-fine strategy improving quality, and Latent-NeRF~\cite{metzer2022latent} performs distillation in latent space leveraging a latent diffusion model~\cite{rombach2021highresolution}. 
These approaches do not specifically target texture generation, which is the focus of this work. More deeply, a crucial drawback of this line of work is that SDS typically requires strong guidance~\cite{ho2021classifierfree} to condition the diffusion model, which can hurt quality and diversity. Moreover, SDS's iterative optimzation process makes synthesis very slow. In contrast, our approach avoids SDS entirely and leverages regular diffusion model sampling in a new, 3D-consistent manner. Earlier works also leverage CLIP~\cite{radford2021clip} for 3D object or texture distillation~\cite{michel2022text2mesh,jain2021dreamfields,khalid2022clipmesh,chen2022tango}, but this performs usually worse than using diffusion models instead.\looseness=-1
\vspace{-6pt}

\paragraph{Concurrent Work} Concurrently with this work, TEXTure~\cite{richardson2023texture} proposes a related approach. 
Like this work, TEXTure performs multiview denoising on a texture map representation.
However,  TEXTure runs an entire generative denoising process in each camera view in sequence, conditioning on the previous views and projecting to the texture map only after full denoising. 
In contrast, in \ourmodel we propose to interleave texture aggregation with denoising steps in different camera views, simultaneously generating the entire output.
This insight significantly reduces view inconsistencies and improves quality in \ourmodel compared to TEXTure, as validated in Sec.~\ref{sec:exp}.
MultiDiffusion~\cite{bar2023multidiffusion} concurrently introduces a method for panorama generation and other controlled image generation tasks, leveraging relatively lower-resolution image diffusion models. Algorithmically, this approach of aggregating different denoising predictions from different image crops is closely related to TexFusion's aggregation from different camera views. However, MultiDiffusion only tackles image synthesis, and is not concerned with any 3D or texture generation at all.\looseness=-1


\section{Background}
\label{sec:Background}

\paragraph{Diffusion Models}
\emph{Diffusion models}~\cite{sohl2015deep,ho2020ddpm,song2020score} model a data distribution $p_{\text{data}}(\rvx)$ via iterative denoising, and are trained with \emph{denoising score matching}~\cite{hyvarinen2005scorematching,lyu2009scorematching,vincent2011,sohl2015deep,ho2020ddpm,song2020score}.
Given samples $\rvx\sim p_{\text{data}}$ and $\boldsymbol{\epsilon} \sim \gN(\mathbf{0}, \mI)$,
a denoiser model $\boldsymbol{\epsilon}_\theta$ parameterized with learnable parameters $\theta$ receives diffused inputs $\rvx_t(\boldsymbol{\epsilon}, t, \rvx)$ and is optimized by minimizing the denoising score matching objective\looseness=-1
\begin{align}
\E_{\rvx \sim p_{\text{data}}, t \sim p_{t}, \boldsymbol{\epsilon} \sim \gN(\mathbf{0}, \mI)} \left[\Vert \boldsymbol{\epsilon} - \boldsymbol{\epsilon}_\theta(\rvx_t; \rvc, t) \Vert_2^2 \right],
\label{eq:diffusionobjective}
\end{align}
where $\rvc$ is optional conditioning information, such as a text prompt, and $p_t$ is a uniform distribution over the diffusion time $t$. The model effectively learns to predict the noise $\boldsymbol{\epsilon}$ that was used to perturb the data (other formulations are possible~\cite{salimans2022progressive,karras2022elucidating}).
Letting $\alpha_{t}$ define a noise schedule, parameterized via 
a diffusion-time $t$, we construct $\rvx_t$ as $\rvx_t = \sqrt{\alpha_t} \rvx + \sqrt{1-\alpha_t} \boldsymbol{\epsilon}, \; \boldsymbol{\epsilon} \sim \gN(\mathbf{0}, \mI)$; this particular formulation corresponds to a variance-preserving schedule~\cite{song2020score}.
The forward diffusion as well as the reverse generation process in diffusion models can be described 
in a continuous-time framework~\cite{song2020score}, but in practice a fixed discretization can be used~\cite{ho2020ddpm}.
The maximum diffusion time is generally chosen such that the input data is entirely perturbed into Gaussian random noise and an iterative generative denoising process 
can be initialized from such Gaussian noise to synthesize novel data.

Classifier-free guidance~\cite{ho2021classifierfree} can be used for improved conditioning. By randomly dropping out the conditioning $\rvc$ during training, we can learn both a conditional and an unconditional model at the same time, and their predictions can be combined to achieve stronger conditioning.



We perform iterative diffusion model sampling via the \textit{Denoising Diffusion Implicit Models (DDIM)} scheme~\cite{song2021denoising}:
\begin{equation}
\begin{split}
    \vx_{i-1} = \sqrt{\alpha_{i-1}}\left(\frac{\vx_i - \sqrt{1 - \alpha_i} \boldsymbol{\epsilon}_\theta^{(t_i)}(\vx_i)}{\sqrt{\alpha_i}}\right) \\
    + \sqrt{1 - \alpha_{i-1} - \sigma_{t_i}^2} \cdot \boldsymbol{\epsilon}_\theta^{({t_i})}(\vx_i) + \sigma_{t_i} \boldsymbol{\epsilon}_{t_i}
\end{split}\label{eqn:ddim}
\end{equation}
with $\boldsymbol{\epsilon}_{t_i}\sim\gN(\mathbf{0}, \mI)$ and $\sigma_{t_i}$ is a variance hyperparameter.
We express obtaining $\vx_{i-1}$ via DDIM sampling as $\vx_{i-1} \sim \vf_\theta^{(t_i)}(\vx_{i-1} | \vx_i)$. See Supp. Material for more details.

\paragraph{Latent Diffusion Models (LDMs) and Stable Diffusion} Instead of directly operating on pixels, LDMs~\cite{rombach2021highresolution} utilize an encoder $\mathcal{E}$ and a decoder $\mathcal{D}$ for translation between images $\xi$ and latents $\rvx\in X$ of a lower spatial dimension. The diffusion process is then defined over the distribution of $X$. \textit{Stable Diffusion} is an LDM trained on the LAION-5B image-text dataset. In addition to text-conditioning, \textit{Stable Diffusion 2.0} permits depth conditioning with a depth map $D$ (\texttt{SD2-depth}). 
This allows detailed control of the shape and configuration of objects in its synthesized images.\looseness=-1

\paragraph{Rendering and Geometry Representation} 
In principle, our method applies to any geometry representation for which a textured surface can be rendered; in practice, our experiments use a surface mesh  $\mathcal{M} = (\mathcal{V}, \mathcal{F})$,  
with vertices $\mathcal{V} = \{v_i\},\ v_i \in \mathbb{R}^3$ and triangular faces $\mathcal{F} = \{f_i\}$ where each $f_i$ is a triplet of vertices. 
Textures are defined in 2D image space in an injective UV parameterization of $\mathcal{M}$, $UV: p \in \mathcal{M} \mapsto (u,v) \in [0,1] ^ 2$.
If needed, this parameterization can be automatically constructed via tools such as \textsc{xatlas}~\cite{theklaatlas2015,xatlas2016}.
We encode textures as multi-channel images discretized at pixels in UV space $\rvz \in \mathbb{R}^{(H \times W, C)}$, notationally collapsing the spatial dimension for simplicity.

We denote the rendering function as $\mathcal{R}(\rvz; \mathcal{M}, C): \rvz \mapsto \rvx $, $\rvx \in \mathbb{R}^{(h \times w, C)}$, which takes as input a mesh $\mathcal{M}$, camera $C$, and texture $\rvz$, and produces as output a rendered image.
The inverse of this function $\mathcal{R}^{-1}(\rvx; \mathcal{M}, C): \rvx \mapsto \rvz$ projects values from camera image-space onto the UV texture map. Notationally, we often omit the dependence on $\mathcal{M}$ and $C$ for brevity.
In this work, we do not model any lighting or shading effects, images are formed by directly projecting textures into screen space (and then decoding, for latent textures).\looseness=-1


\section{Texture Sampling with 2D Diffusion Models} 

\begin{figure}[h]
    \centering
    \setlength{\tabcolsep}{0pt}
    \vspace{-0.15cm}
    \includegraphics[trim=1.0cm 1.5cm 1.0cm 1.5cm,width=0.8\linewidth]{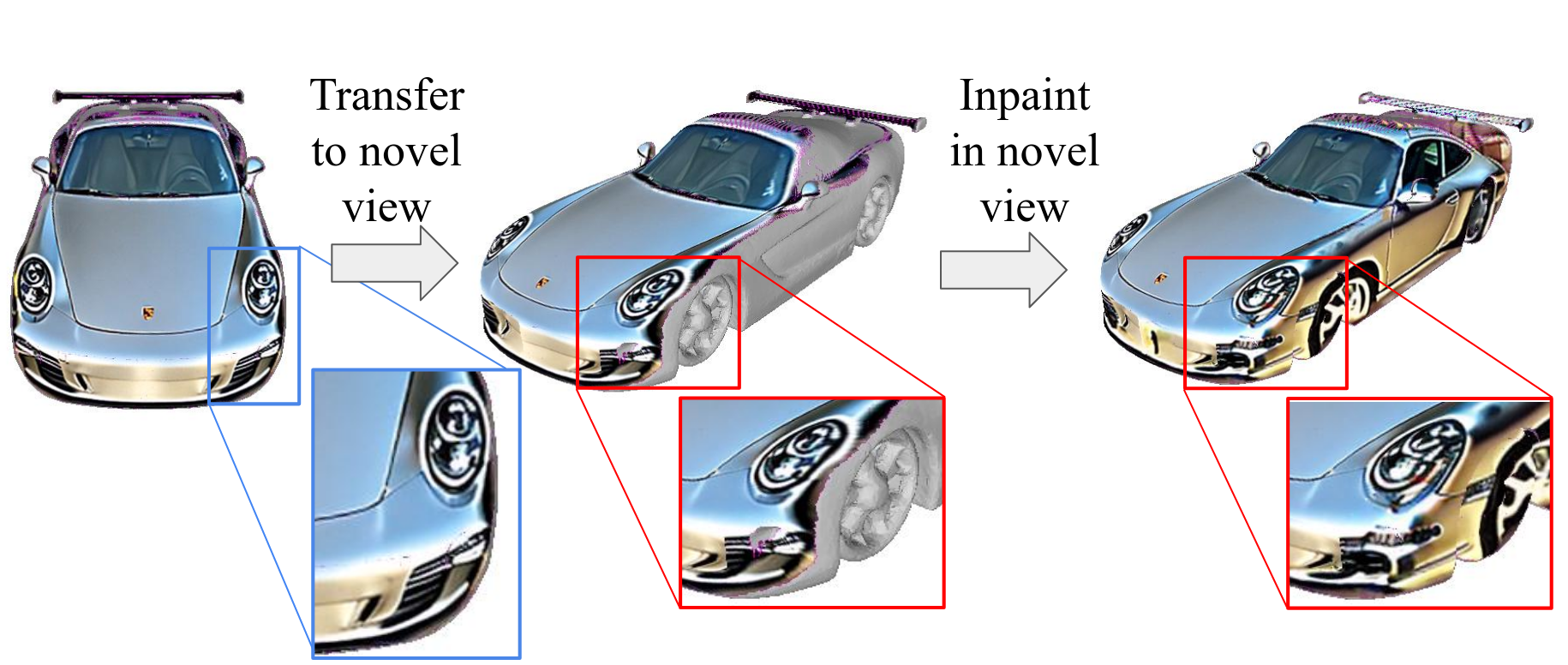}
    \caption{\small Illustration of irreconcilable mistakes in early views impacting denoised results in later views (images sampled from TEXTure). While the highlighted area appears natural in the first view, it does not match the geometry when viewed from a different angle, thereby creating poorly denoised results when the second view is inpainted.}
    \label{fig:inconsistency}
    \vspace{-0.4cm}
\end{figure}

Given a mesh geometry $\mathcal{M}$, a text prompt $y$, and a geometry (depth) conditioned image diffusion model $\theta$, how could one sample a complete surface texture? 
Assuming access to the rendering $\mathcal{R}$ and inverse rendering $\mathcal{R}^{-1}$ functions defined above, perhaps the most naive approach is to compute a set of $\{C_1, .., C_N\}$ camera views that envelopes the surface, render the depth map $d_n$ in each view, sample images from the image diffusion model with depth conditioning, and then back-project these images to the mesh surface (\eg as done in \cite{dreamtexture2022}). 
However, image diffusion models in each view have no information about the generated results in other views, thus there is no coherency in the contents generated in each view. As an alternative, one may define a canonical order of camera views, and autoregressively condition image sampling in the subsequent views on previously sampled regions (as done in \cite{richardson2023texture, metzer2022latent}). However, for most geometries of interest (i.e. not a plane), a single camera can not observe the entirety of the geometry. Consequently, images synthesized early in the sequence could produce errors that are not reconcilable with the geometry that is observed later in the sequence (see Fig.~\ref{fig:inconsistency}). Thus, it is desirable for the image sampling distribution $p(\cdot|d_i, y)$ in each camera view to be conditioned on that of every other camera view.

\subsection{Sequential Interlaced Multiview Sampler} \label{sec:SIMS}
Leveraging the sequential nature of the denoising process, we can interlace the synchronization of content across views with the denoising steps within each view to achieve coherency over the entire shape. 
Suppose that at step $i$ of the denoising process, we have a set of partially denoised images $\{\rvx_{i, n}\}_{n=1}^N = \{\rvx_{i, 1} ... \rvx_{i, N}\}$. Our goal is to sample $\{\rvx_{i-1, n}\}_{n=1}^N = \rvx_{i-1, 1} ... \rvx_{i-1, N}$ that is 3D-consistent, i.e., two pixels in $\rvx_{i-1, a}, \rvx_{i-1, b}$ that project to the same point in 3D should have the same value. 
Taking inspiration from the autoregressive modeling literature, we sample 
the joint distribution via first decomposing it into a product of conditionals, and approximating each term in the product by using the diffusion model to denoise the renders of a dynamically updated latent texture map. Specifically, we first initialize an initial latent texture map $\rvz_{T} \sim \mathcal{N}(\mathbf{0}, \boldsymbol{I})$, ($T$ demarks the first step of the diffusion trajectory). 
Then, suppose that we have a 3D consistent latent texture map $\rvz_{i}$, we decompose the joint distribution as follows (conditioning on depth and prompt omitted for space):
\begin{equation}
\begin{split}
    p_\theta( \{\rvx_{i-1, j}\}_{j=1}^N | \rvz_{i}) = p_\theta(\rvx_{i-1,1} | \rvz_{i}) \times \\
    \prod_{n=2}^N p_\theta(\rvx_{i-1, n} | \{\rvx_{i-1, j}\}_{j=1}^{n-1},  \rvz_{i}) \end{split}\label{eqn:seq_sampling}
\end{equation}
We can compute the first term by first rendering $\rvz_i$ into $\rvx_{i, 1}' = \mathcal{R}(\rvz_i; C_1)$. Eqn.~\ref{eqn:ddim} can now be applied to $\rvx_{i, 1}'$ to draw a latent image at the next time step:
\begin{equation}
    \rvx_{i-1, 1} \sim f_\theta^{(t_i)}(\rvx_{i-1, 1}| \rvx_{i, 1}' = \mathcal{R}(\rvz_i; C_1)). \label{eqn:parallel_sample}
\end{equation}
Later terms in Eq.~\ref{eqn:seq_sampling} additionally depend on the result of previous denoising steps.
We again model this dependency through the latent texture map. For each view starting at $n=1$, we inverse render $\rvx_{i-1, n}$ into texture space to obtain $\rvz_{i-1, n}' = \mathcal{R}^{-1}(\rvx_{i-1, n})$, and update the pixels of $\rvz_{i-1, n-1}$ that are newly visible in $\rvz_{i-1, n}'$ to obtain $\rvz_{i-1, n}$ (See Sec. \ref{sec:aggregating_latents} for details).
Then, in the $n+1$ iteration, since $\rvz_{i-1, n}$ contain regions at two noise scales (unseen regions are at noise scale $\sigma_i$, while visited regions are at noise scale $\sigma_{i-1}$), we add appropriate 3D consistent noise to the visited regions of $\rvz_{i-1, n}$ to match the noise scale of step $i$ before rendering it as input to $f_\theta$. Letting $M_{i,n}$ represent the mask for visited regions and $\epsilon_i \sim \mathcal{N}(\mathbf{0}, \boldsymbol{I})$, we can write the sampling procedure for $\rvx_{i-1, n}$ as:
\begin{equation}
\begin{split}
    \rvz_{i, n} &= M_{i,n} \odot \left(\sqrt{\frac{\alpha_{i-1}}{\alpha_i}}  \rvz_{i-1, n-1} + \sigma_i \epsilon_i \right) \\ & + \left( \mathbf{1} - M_{i,n}\right) \odot \rvz_i \\
    \rvx_{i-1, n} &\sim f_\theta^{(t_i)}(\rvx_{i-1, n}| \rvx_{i, n}' = \mathcal{R}(\rvz_{i,n}; C_n))
\end{split}\label{eqn:SIMS}
\end{equation} 
By iteratively applying Eqn.~\ref{eqn:SIMS}, we obtain a sequence of 3D consistent images $\{\rvx_{i-1, n}\}_{n=1}^N$ and a texture map $\rvz_{i-1, n}$ that has been aggregated from these images. We can then decrement $i$ and repeat this process in the next time step. 

We name this approach Sequential Interlaced Multiview Sampler (SIMS). SIMS communicates the denoising direction of previous views to the current view and resolves overlaps between views by the order of aggregation. It ameliorates inconsistent predictions while circumventing performance degradation due to the averaging of latent predictions during parallel aggregation. In the single-view case, SIMS is equivalent to standard DDIM sampling. A complete algorithm for SIMS can be found in the appendix.

\subsection{The \ourmodel Algorithm}
In Sec.~\ref{sec:SIMS}, we have presented a generic algorithm for sampling 3D consistent multi-view images and texture maps using 2D diffusion models. We now present a concrete algorithm, \ourmodel, that uses SIMS to texture 3D meshes using \texttt{SD2-depth}~\cite{rombach2021highresolution} as the diffusion model. An illustrative overview of \ourmodel can be found in Fig.~\ref{fig:overview_figure}. 

As \texttt{SD2-depth} is a latent diffusion model, we apply SIMS in the latent space: $\rvx$ and $\rvz$ represent latent images and latent texture maps respectively, and Section~\ref{sec:distillation} will describe a final distillation post-process to color space. 
In this section, we address several challenges specific to using LDMs with SIMS, and detail our design choices to tackle these challenges. 
We find that a canonical set of cameras works well for most objects, but cameras are tailored to specific objects to improve resolution and ameliorate occlusion. We further illustrate a technique for obtaining high-quality results by operating SIMS in a cascade of increasing resolutions. 
\vspace{-0.3cm}

\subsubsection{(Inverse) Rendering of Latent Textures}
We use \textsc{nvdiffrast} \cite{Laine2020diffrast} to efficiently render textured meshes via rasterization, as described in Sec.~\ref{sec:Background}.
Our implementation sets the rendered image size $h=w=64$ to match Stable Diffusion's UNet, with latent vectors of dimension $D=4$.
The texture image dimensions $H$ and $W$ are chosen based on the surface area of the mesh relative to its diameter (detailed in Sec.~\ref{sec:geometry_and_camera}).

For each non-background pixel $s$ in a rendered image $\rvx_{i,\cdot}'$, rasterization gives a corresponding location on the texture image via the UV map $(u,v) = UV(p)$, and we retrieve the value at the nearest texture pixel.
This texture value is latent, and no shading or lighting is applied.
In other settings, texture data is often rendered with bilinear filtering and mipmaps to improve image quality and reduce aliasing, but in this setting, we found it essential to \emph{avoid} such techniques. We experimentally ablate texturing approaches in the supplementary. 
The issue is that early in the diffusion process, i.e. when $t_i \ll 0$, $\rvz_{i}$ and $\rvx_{i, \cdot}'$ are dominated by noise, but interpolation and mipmapping change the variance in the pixels $\rvx_{i, \cdot}$, thereby moving $\rvx_{i, \cdot}'$ out of the training distribution of \smash{$\epsilon_\theta^{(t_i)}$}.
Instead, we retrieve only the nearest texture pixels for diffusion, and resolve aliasing and interpolation via a simple distillation post-process (Sec.~\ref{sec:distillation}). 
For each background pixel in rendered image $\rvx_{i,\cdot}'$, we apply a Gaussian noise of standard deviation $\sigma_i$, such that the background matches the marginal distribution at diffusion step $t_i$. This ensures that the diffusion model $f_\theta^{(t_i)}$ focuses solely on the foreground pixels of $\rvx_{i,\cdot}'$.

Note that in this setting a rendered image is simply a selection of pixels from the texture maps, and thus we can leverage backpropagation to easily implement the inverse rendering function $\mathcal{R}^{-1}$.
Additionally, forward rasterization yields a depth map of the scene, which we use as a conditioning input to the diffusion model.

\begin{figure}[h]
    \centering
    \setlength{\tabcolsep}{0pt}

    {\small
    \begin{tabular}{c c}
        \includegraphics[width=0.09\textwidth,trim=0 0 0 0,clip]{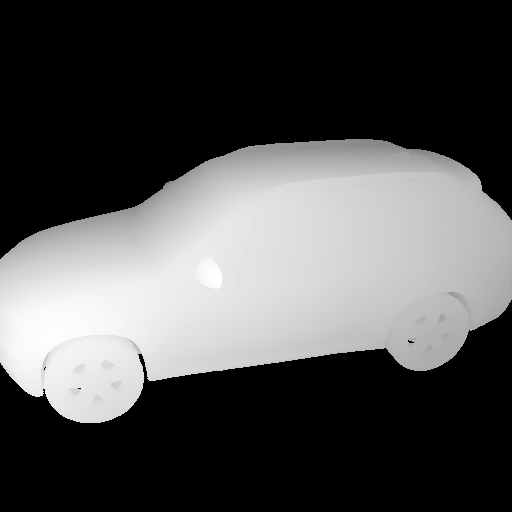}
        \includegraphics[width=0.09\textwidth,trim=0 0 0 0,clip]{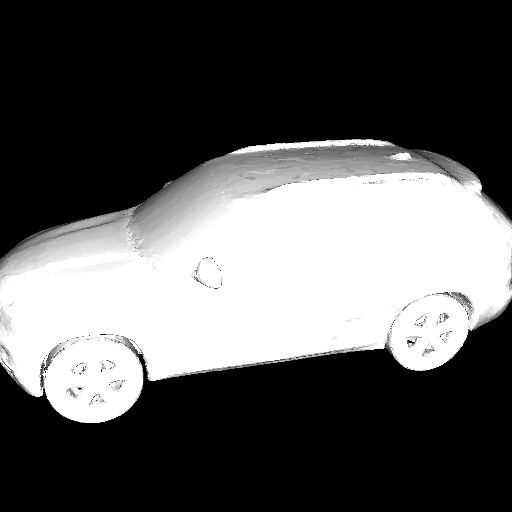} &
         \includegraphics[width=0.203\textwidth,trim=0 0 0 0,clip]{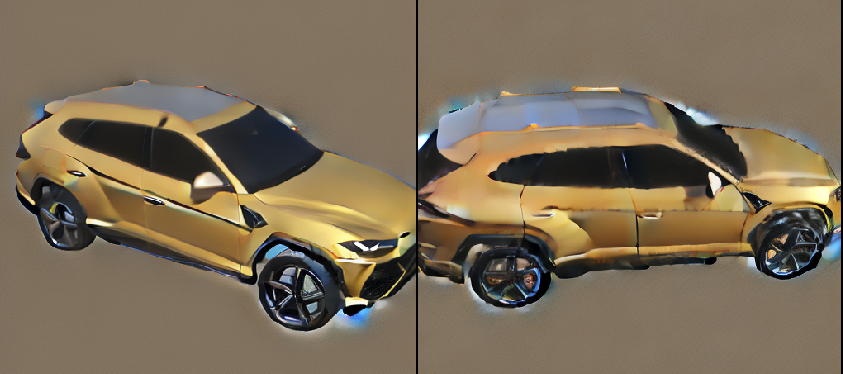}
    \end{tabular}
    }
    
    \caption{\small Left: Depth map for conditioning \texttt{SD2-depth}, and quality image computed from screen space derivatives. Right: Output of SD2 decoder in two views using renders of the latent texture map as input. Note how the horizontal line across the doors changes appearance from one view to another.}
    \label{fig:quality}
    \vspace{-0.4cm}
\end{figure}

\subsubsection{Aggregating Latent Textures.}\label{sec:aggregating_latents} Recall from Sec.~\ref{sec:SIMS}, before iterating through cameras $\{C_1, ..., C_N\}$, we first initialize a latent texture map $\rvz_{i-1, 0} = \rvz_{i}$ and render latent image $\rvx_{i, 1}' = \mathcal{R}(\rvz_{i-1, 0})$. Then, in each step (iterating through cameras), we obtain $\rvx_{i-1, n}$ from $f_\theta^{(t_i)}(\rvx_{i,n}')$, and inverse render it to get $\rvz_{i-1, n}' = \mathcal{R}^{-1}(\rvx_{i-1, n})$. Finally, we need to aggregate the partial texture map $\rvz_{i-1, n}'$ with $\rvz_{i-1, n-1}$ to obtain a partially updated texture map $\rvz_{i-1, n}$. We perform this aggregation step based on a simple heuristic that the value of each pixel $(u,v)$ in $\rvz$ should be determined by the camera that has the most ``direct and up-close" view to its corresponding point on the mesh. We measure view quality using image-space derivatives - the amount of change in UV coordinates per infinitesimal change in the image coordinate. This quantity is commonly used for determining mipmapping resolution and anti-aliasing, and it can be efficiently computed by nvdiffrast when rasterizing $\rvz_{i, n-1}$. For each pixel location $(p,q)$ in $\rvx_{i-1, n}$, we compute the negative Jacobian magnitude as $-|\frac{\partial u}{\partial p} \cdot \frac{\partial v}{\partial q} - \frac{\partial u}{\partial q} \cdot \frac{\partial v}{\partial p}|$, and inverse render it to the texture space, which we denote as $Q_{i,n}$. Higher values of $Q_{i,n}(u,v)$ means that camera $n$ has a better view of $(u,v)$ (see Fig.~\ref{fig:quality}). 
In addition, we compute $M_{i,n} = \mathcal{R}^{-1}(\boldsymbol{I})$, such that $M_{i,n}(u,v)$ represents the number of pixels in $\rvx_{i,n}'$ that received value from $\rvz_{i}(u,v)$. 

While iterating through cameras $1$ through $N$, we maintain mask $M_i$ (initialized to $0$), which is used to track which pixels of $\rvz_{i,n}$ have been seen by cameras up to $n$, and quality buffer $Q_i$ (initialized to $-\inf$), which is used to track the highest non-zero value of $Q_{i,n}$ at each $(u,v)$. The value at $(u,v)$ of $\rvz_{i-1, n}$ is determined as:
\begin{equation}
    \rvz_{i-1, n}(u,v) = \begin{cases}
        \frac{\rvz_{i-1, n}'(u,v)}{M_{i, n}(u, v)} &  \begin{split}M_{i, n}(u, v) > 0 \text{ , and} \\ Q_{i, n}(u,v) > Q_i(u,v)\end{split}\\
        \rvz_{i-1, n-1}(u,v)                 & \text{otherwise.} \label{eqn:aggregation}
\end{cases}
\end{equation}
$M_{i}$ and $Q_i$ are then updated pixelwise as $M_i = \mathit{\textrm{min}}(M_{i, n} + M_i, 1)$ and $Q_i = \max{(Q_{i,n}, Q_i)}$ (min and max are applied element-wise).
We use Eqn.~\ref{eqn:aggregation} in conjunction with Eqn.~\ref{eqn:SIMS} in SIMS to perform denoising of images in the sequence of camera views. 

We further note that a side effect of using nearest pixel filtering during texture sampling in SIMS is aliasing. 
When the image space derivative Jacobian is much higher than 1 (low quality), neighboring pixels in screen space will have gaps when mapped to uv space. This results in incomplete filling of the latent texture map.
Conversely, when the image space derivative Jacobian magnitude is much smaller than 1 (high quality), multiple screen space pixels $(p,q$ will map to the same $(u,v)$, creating uniform blocks in $\rvx$. Similar to interpolated latents, these blocks are particularly detrimental early during the diffusion process, as they cannot be correctly denoised by $\epsilon_\theta$ which is expecting high spatial variance.  Our quality-based aggregation overwrites low-quality regions in the latent texture when higher quality views are available. Thus we can set the resolution of the texture map such that the image space derivatives in each view do not fall below $1$ (i.e. $\max Q_{i,n} < -1$) to prevent aliasing of the second type, and rely on cameras with better views to fix aliasing of the first type.

\vspace{-4mm}
\subsubsection{From Latent to RGB Textures} 
\label{sec:distillation}
So far, we have described how to use SIMS to produce a set of 3D consistent latent images and latent texture maps, but have yet to describe how to translate this into a RGB texture map that can be rendered for viewing. Towards this, we experimented with multiple approaches and found performing multi-view distillation of decoded latent images $\{\rvx_{0, n}\}_{n=1}^N$ with a neural color field to be most performant. 
Specifically, we use the decoder $\mathcal{D}$ of Stable Diffusion to decode latent multi-view images $\{\rvx_{0, n}\}_{n=1}^N$ into RGB multi-view images $\{\xi_{n} = \mathcal{D}(\rvx_{0, n})\}_{n=1}^N$. Notably, decoding with $\mathcal{D}$ introduce inconsistencies such that $\{\xi_{n}\}_{n=1}^N$ is not 3D consistent even when $\{\rvx_{0, n}\}_{n=1}^N$ is 3D consistent (see Fig.~\ref{fig:quality} for example). In stable Diffusion, each latent vector (pixel of $\rvx$) needs to carry the information of a $8 \times 8$ patch of RGB pixels. Thus, the value of the latent vector encodes both color values and spatial patterns. We therefore cannot expect their decoded results to be equivariant to perspective transformations. 
To address this problem, we leverage a neural color field optimized with appearance-based losses to smooth out inconsistencies.
Since we know the ground-truth camera poses, we can directly obtain the 3D spatial coordinates $\{\mathit{xyz}_{n}\}_{n=1}^N$ of all pixels of $\{\xi_{n}\}_{n=1}^N$ by projecting pixels to the mesh surface. Background pixels that do not intersect any surface are discarded. Following \cite{lin2022magic3d}, we use a multi-resolution hash encoding based on instantNGP \cite{muller2022instantngp} along with a shallow MLP $f_\phi$ to parameterize a function from 3D spatial coordinates to RGB values for each sample $\mathit{rgb} = f_\phi(\mathit{hash}(\mathit{xyz}))$. We then distill multi-view images $\{\xi_{n}\}_{n=1}^N$ into this parametric function via optimization. 
Since our goal is to export $\phi$ into a texture map, we do not use any view-dependent parameterization. To reconcile inconsistencies, we use both a standard $L2$ loss and a VGG-based perceptual loss, applied between $f_\phi$ and $\{\xi_{n}\}_{n=1}^N$, to train $\phi$. We use Adam with a learning rate of $0.01$, and optimization of $\phi$ converges within $100$ iterations. After optimization, we compute the spatial coordinate of the centers of each pixel in a high-resolution texture map, and query $f_\phi$ to predict RGB values for the texture map.

\vspace{-5pt}
\subsubsection{Geometry Processing, Camera, and Multi-resolution Refinement}\label{sec:geometry_and_camera}
We normalize $\mathcal{M}$ such that it fits inside a cube of side length 1, and center it at the origin. 
Perspective cameras are placed facing the origin, and their FOV is adjusted to fit the object. Detailed parameters can be found in the appendix. 
As the diffusion model relies on context captured in $\rvx_{i, \cdot}$ to perform denoising, camera views that are too small w.r.t the size of the object often result in content drift - the texture in distant areas of the mesh can have content that is semantically inconsistent. This can be seen as our version of the Janus face problem known to Dreamfusion and similar approaches \cite{poole2022dreamfusion, lin2022magic3d, wang2022scorejacobian}. 
Our solution is to perform two rounds of \ourmodel at different resolution scales to obtain high-resolution textures while retaining semantic consistency.

Specifically, we first run \ourmodel using cameras that cover the entire object, and a low-resolution latent texture map that is suitable for these cameras. We do not run view distillation in this first round as we are only interested in the latent texture map. We denote the denoised latent texture map from this step as $\rvz_{0, lr}$. We then use a second set of cameras with a narrower field of view; these cameras are also more numerous to still cover the full object. We determine a new texture map resolution using the square of the ratio of the tangent of the old FOV over the new FOV - corresponding to the relative change in surface area covered by the camera before and after the change in FOV. We then up-sample $\rvz_{0, lr}$ with nearest neighbor filtering to this higher resolution, and stochastically encode it to a partially noised state (\eg $T=500$ in the diffusion model time schedule). The second round of \ourmodel uses these new cameras, and initializes SIMS with the partially noised latent texture map. The multi-view images produced by the second round of SIMS is used to produce the final output texture map via neural color field distillation. 

\section{Experiments} \label{sec:exp}
We apply \ourmodel on various geometries and text prompts to evaluate its ability to produce high quality, natural, and 3D consistent textures. We focus our experimental comparisons on TEXTure \cite{richardson2023texture}, a text-driven texture generation method that also uses \texttt{SD2-depth}. We choose TEXTure as the baseline because (1) it represents the current state-of-the-art for language-conditioned texture generation, (2) it uses the same \texttt{SD2-depth} model, which can be prompted for a wide variety of content, and (3) it is concurrent to our work. In the supplementary materials, we further compare TexFusion to SDS-based text-driven texture distillation~\cite{poole2022dreamfusion,metzer2022latent}, leveraging the Stable Diffusion model and show that TexFusion achieves superior results in terms of quality and speed.

\vspace{-2mm}
\paragraph{Dataset} We collect 35 meshes of various content, and write 2-4 prompts for each mesh. In total, we evaluate each texture synthesis method on 86 mesh-text pairs. More details of this dataset are in the supplementary materials.

\begin{figure}[b]
    \centering
    \setlength{\tabcolsep}{0pt}
    {\small
        \includegraphics[trim={1.5cm 9.5cm 1.5cm 2.8cm},clip,width=0.87\linewidth]{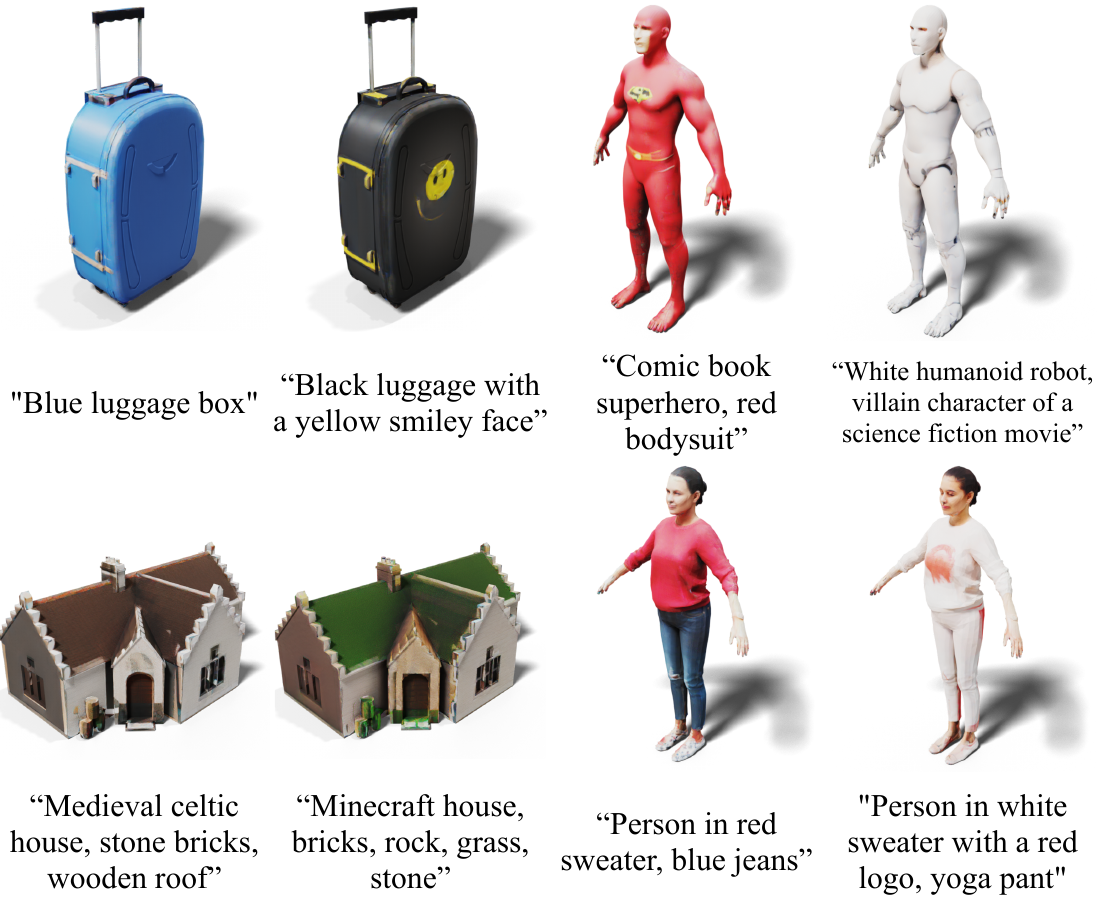} 
    }
    \vspace{-0.3cm}
    \caption{\small More TexFusion text-conditioned texturing results.
    \vspace{-0.3cm}
    }
    \label{fig:more_results}
\end{figure}

\subsection{Qualitative Comparisons}


\begin{figure*}[h!]
    \centering
    \includegraphics[trim={0.5cm 20.3cm 1cm 1.2cm},clip,width=\linewidth]{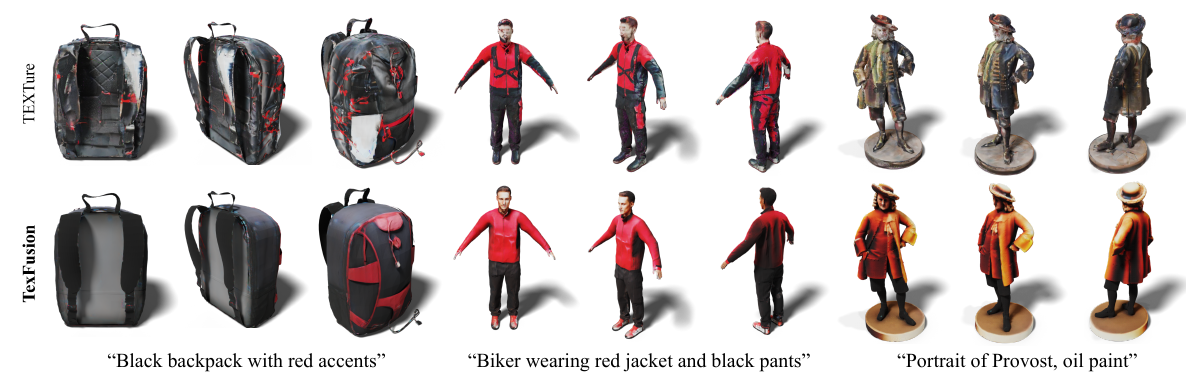}
    \includegraphics[trim={0.5cm 20.5cm 1cm 1.5cm},clip,width=0.8\linewidth]{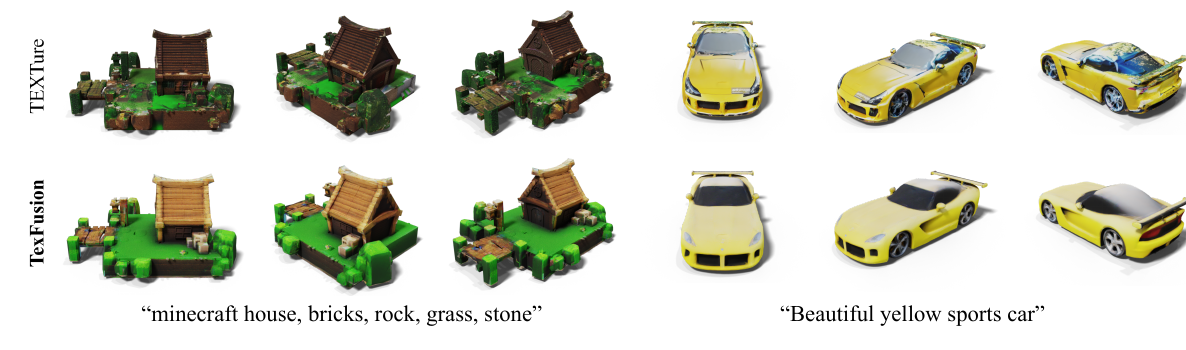}
    \vspace*{-2em}
    \caption{\small Visual comparison of textures generated by TEXTure~\cite{richardson2023texture} and TexFusion.
    \vspace*{-2em}
    }
    \label{fig:comparison}
\end{figure*} 

We visualize textures produced by \ourmodel on multiple geometries and prompts, and compare to state-of-the-art baselines in Figure~\ref{fig:comparison}.
We render each object in 3 surrounding views to allow better examination of 3D consistency of the produced texture. 
We use Blender's \texttt{Cycles} renderer with a studio light-setup. Textures are applied as base color to a diffuse material.
Additional visualizations, including videos showing 360 pans of all objects presented in the paper, and renders of our results using only texture color, can be found in the supplementary materials.

In terms of local 3D consistency (consistency in neighborhoods on the surface of the mesh), textures produced by \ourmodel are locally consistent - there are no visible seam lines or stitching artifacts. In contrast, we often find severe artifacts when viewing the top and back sides of TEXTure's outputs. These artifacts are most noticeable when a clean color is expected, such as when texturing vehicles. 
In terms of global consistency (semantic coherency of the entire texture, e.g. exactly 2 eyes and 1 nose to a face), TEXTure performs poorly and suffers from problems similar to DreamFusion's Janus face problem~\cite{poole2022dreamfusion}: as the 2D diffusion model captures context of the object through its own camera view, it is not aware of the appearance of the opposite side of the object. This problem is ameliorated in \ourmodel due to frequent communication between views in SIMS.\looseness=-1

There are noticeably more baked specular highlights and shadows in textures generated by TEXTure. These effects are physically inaccurate as they conflict with the lighting effects simulated by the renderer.  In contrast, \ourmodel produces textures that are smoothly illuminated. 
We hypothesize that this is due to interlacing aggregations in SIMS, which removes view-dependent effects as they arise in the sampling process. We provide additional visualizations of \ourmodel in Fig.~\ref{fig:more_results} to showcase how text prompting is effective at controlling the generation of textures, thereby producing varying appearances using the same mesh.

\paragraph{Runtime} \ourmodel takes approximately 3 minutes on a machine with a single GPU to sample one texture. We are slightly faster than TEXTure (5 min.) \cite{richardson2023texture}, since we only need to optimize a color field once after sampling is complete. We are also an order of magnitude faster than methods that rely on SDS loss which report upwards of 40 minutes \cite{poole2022dreamfusion,metzer2022latent, lin2022magic3d}. Additional runtime comparisons are in the appendix.\looseness=-1



\subsection{Quantitative Comparisons}
\begin{table}
\small
\centering
\setlength{\tabcolsep}{3pt}
\scalebox{0.82}{
\begin{tabular}{l c c c c c} 
\toprule
\multirow{2}{*}{Method} & \multirow{2}{*}{FID ($\downarrow$)} & \multicolumn{4}{c}{User study (\%)} \\
 & & \begin{tabular}{c}  Natural \\ Color ($\uparrow$) \end{tabular} & \begin{tabular}{c}  More \\ Detailed ($\uparrow$) \end{tabular} & \begin{tabular}{c}  Less \\ Artifacts ($\uparrow$) \end{tabular}  & \begin{tabular}{c}  Align with \\ Prompt ($\uparrow$) \end{tabular}  \\
\midrule
TEXTure    & 79.47          &  24.42   &  $\textbf{65.12}$  & 31.40 & 43.02\\
$\textbf{\ourmodel}$ & $\textbf{59.78}$ & $\textbf{75.58}$  & 34.88 & $\textbf{68.60}$ & $\textbf{56.98}$ \\
\bottomrule
\end{tabular}
}
\caption{ \small Quantitative and qualitative comparisons between TEXTure and \ourmodel{}. FID is computed w.r.t. a set of images synthesized by \texttt{SD2-depth}.
}
\vspace{-3mm}
\label{tb:fid}
\end{table}
It is inherently difficult to quantitatively evaluate the quality of text and geometry-conditioned texture generation as one must take alignment with the conditioning inputs into account. We rely on both automated and human evaluation to gauge the quality of the synthesized textures.
\paragraph{FID} Since \texttt{SD2-depth} can produce high-quality images that match the structure of the conditioning geometry, we sample it to create a proxy ground truth set. Furthermore, as both TEXTure and \ourmodel use \texttt{SD2-depth} as 2D prior, samples from it serve as an upper bound to the quality and diversity of the appearance of textures: We render the depth map of all meshes in 8 canonical views, and condition \texttt{SD2-depth} on both these depth maps and our text prompts to generate a ground truth set. We also render textures generated by both methods in these same 8 views, and compute the Fr\'{e}chet Inception Distance (FID) between images rendered from the textures and the ground truth set. FID measures the overlap between two sets of images based on their similarity in a learned feature space.
We white out the background regions in rendered images and ground truth images to focus the comparisons to the textured object.
We present the results in Tab.~\ref{tb:fid}. FID scores indicate that textures synthesized by \ourmodel are closer in appearance to the ground truth set than those from TEXTure. The absolute FID values are likely high due to the relatively small set of images in each dataset. 
\vspace{-3mm}
\paragraph{User study}
We conducted a user study to measure the overall quality of the generated textures. We use all mesh and text prompts in our dataset, and render each textured mesh into a video showing the results from a $360^\circ$ rotating view. We present videos from TEXTure and \ourmodel side by side in random left-right order, to avoid bias. The user study was conducted on Amazon Mechanical Turk.
Each participant is asked to select the preferred result under four criteria: natural color (non-saturation), details, cleanliness, and alignment with the text prompt. 
To avoid randomness in the answers, we let 3 participants answer the same question and determine the choice by max-voting, a question screenshot is provided in the supplement.
We provide the user study results in Tab.~\ref{tb:fid}. Although not deemed as detailed as TEXTure, our results are overall preferred by the users in terms of better aligning with the provided text prompt, and having more natural color, and fewer artifacts/flaws.  

\subsection{Improving Texture Details}\label{sec:improving_details}
\begin{figure}[t]

    \centering
    \vspace{-0.15cm}
    \setlength{\tabcolsep}{1pt}
    \includegraphics[trim={1.5cm 12.2cm 1.5cm 2.5cm},clip,width=0.95\linewidth]{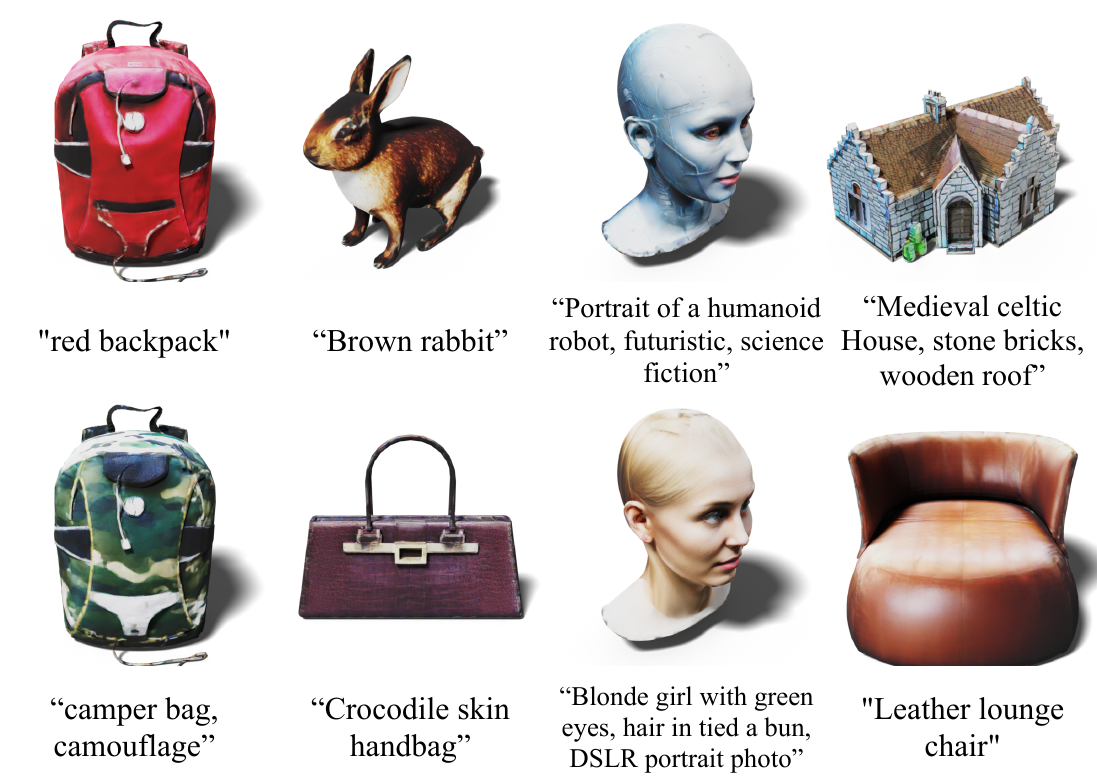}
    
    \caption{\small TexFusion + non-stochastic DDIM sampling ($\sigma_{t_i}=0$). This setting emphasizes textural details, such as the leather (bottom, left), roof shingles (top, right)
    }
    \label{fig:even_more_results2}
\end{figure}

\begin{figure}[t]

    \centering
    \vspace{-0.15cm}
    \setlength{\tabcolsep}{1pt}
    \includegraphics[trim={3cm 1.3cm 11.5cm 12.5cm},clip,width=0.95\linewidth]{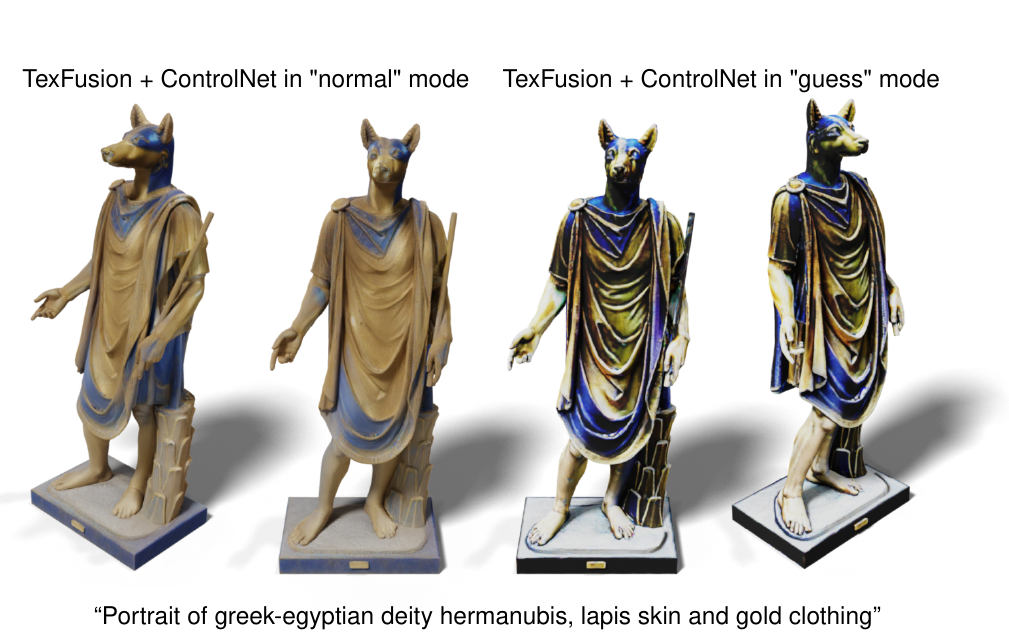}
    
    \caption{\small Left: TexFusion + ControlNet in ``normal mode"; right: TexFusion + ControlNet in ``guess mode".
    \vspace{-3mm}
    }
    \label{fig:cnet_results}
\end{figure}

\begin{figure}[t]

    \centering
    \vspace{-0.15cm}
    \setlength{\tabcolsep}{1pt}
    \includegraphics[trim={1.5cm 20cm 1.5cm 1.8cm},clip,width=0.9\linewidth]{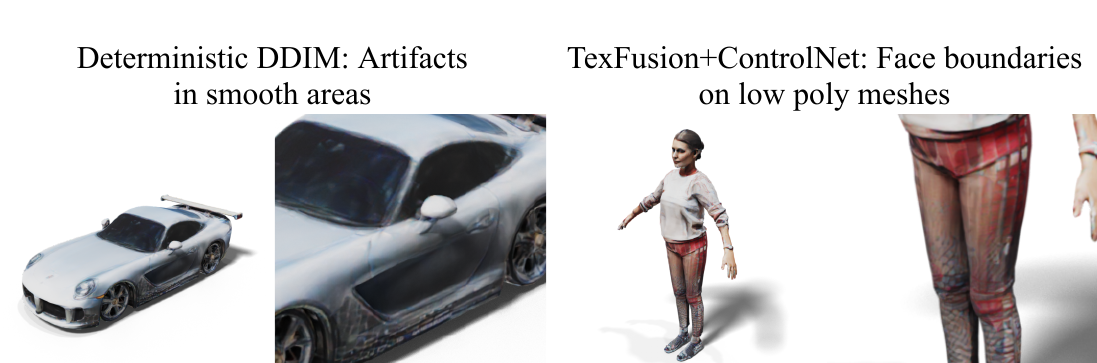}
    
    \caption{\small Failure modes exhibited by (left) TexFusion + non-stochastic DDIM sampling and (right) TexFusion + ControlNet.
    \vspace{-3mm}
    }
    \label{fig:failed_results}
\end{figure}

Texfusion can produce more texture details with adjusted hyperparameters and diffusion backbones. 
First, we find that using the non-stochastic version of DDIM ($\eta= 0$) adds more materialistic details to the textures on smooth/low-poly geometries.
We showcase some examples with particularly large improvements in Fig.~\ref{fig:even_more_results2}. 
Second, we explore the use of ControlNet\cite{zhang2023adding}, which can be easily substituted as the LDM backbone for our method without any additional changes. We find that its high-resolution depth conditioning allows TexFusion to capture fine-grained geometric details in the input mesh. In Fig.~\ref{fig:cnet_results}, we further compare TexFusion + ControlNet in ``normal mode" (apply classifier-free guidance to text prompt only) and ControlNet's ``guess mode" (apply classifier-free guidance to both text and depth) on meshes with fine geometric details. TexFusion produces high-contrast textures with the appearance of strong lighting under ``guess mode", and realistic textures with smooth lighting in ``normal mode". 
{These modifications improve details at a cost of reduced robustness. The non-stochastic DDIM setting may create artifacts when a smooth and clean texture is desired.
On the other hand, the increase in depth resolution makes TexFusion+ControlNet susceptible to capturing face boundaries on low-poly meshes. We provide visualizations of these failure cases in Fig.~\ref{fig:failed_results}. 
Nonetheless, these modifications offer further dimensions of control that can be explored by practitioners. }

\section{Conclusion and Limitations}
We presented \textit{TexFusion}, a novel approach to text-driven texture synthesis for given 3D geometries, using only large-scale text-guided image diffusion models. TexFusion leverages the latent diffusion model \textit{Stable Diffusion} and relies on a new 3D-consistent diffusion model sampling scheme that runs a 2D denoiser network in different views of the object and aggregates the predictions in a latent texture map after each denoising iteration. We find that our method can efficiently generate diverse, high-quality, and globally coherent textures and offers detailed control through text conditioning. Limitations of our approach include a not-yet-ideal sharpness and that texture generation is not real-time. Future work can address these issues; for instance, it would be interesting to leverage the recent literature on faster diffusion model samplers~\cite{dockhorn2022genie,zhang2022fast,lu2022dpm,liu2022pseudo}.

\paragraph{Concurrent Works Since Submission}
Since the time of writing this paper, new works has appeared in the text to texture space \cite{chen2023text2tex, Tang2023TextguidedHC}. They improve upon TEXTure in aspects such as adding a refinement procedure to automatically fix low quality areas from the initial texturing process \cite{chen2023text2tex}, and using images sampled from a text-to-image model and per-prompt finetuning to provide stronger conditioning \cite{Tang2023TextguidedHC}. \ourmodel is methodologically distinct from these methods which are based-on TEXTure. Improvements proposed in these works could be combined with \ourmodel in future work. 

\vspace{-0.3cm}
\paragraph{Acknowledgements} We would like to thank Jun Gao for the helpful discussions during the project. Tianshi Cao acknowledges additional income from Vector Scholarships in Artificial Intelligence, which are not in direct support of this work.

{\small
\bibliographystyle{ieee_fullname}
\bibliography{egbib}
}
\clearpage 
\newpage
\section{Algorithm Details}\label{sec:camera}

\begin{figure}[t!]
    \centering
    \includegraphics[width=0.6\linewidth]{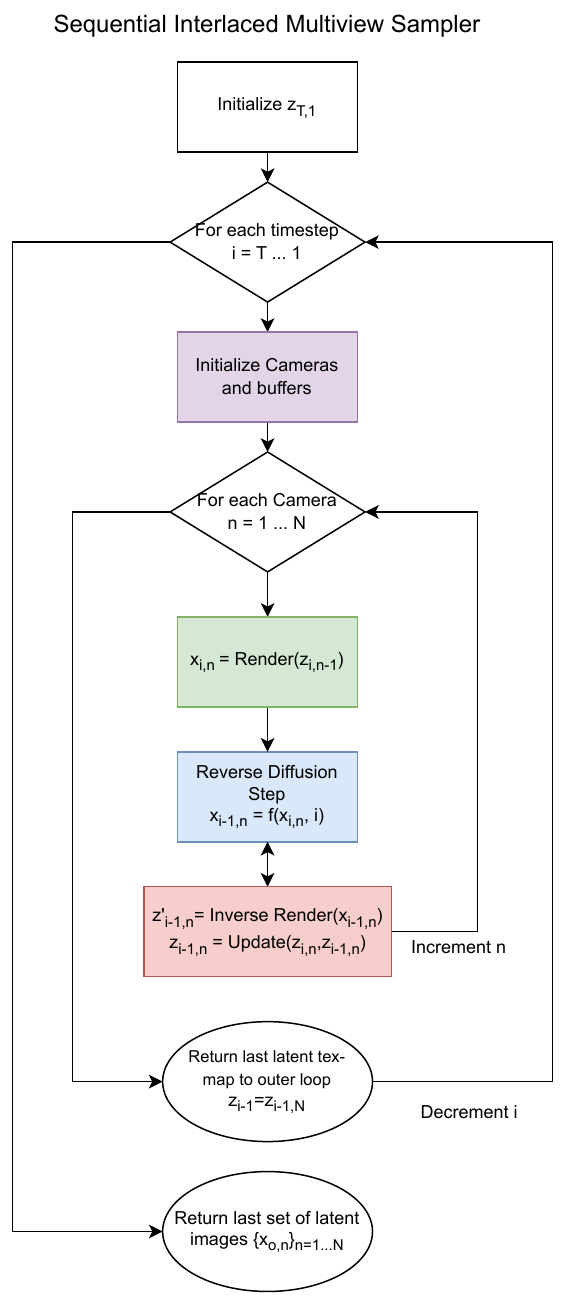}
    \caption{\small Simplified block diagram of SIMS.
    \vspace{-3mm}
    }
    \label{fig:diagram}
\end{figure}

\paragraph{Algorithm}
We present a full itinerary for the Sequential Interlaced Multiview Sampler in  Algorithm~\ref{alg:sims} and a simplified block diagram in Fig.\ref{fig:diagram}. The symbol $\boldsymbol{I}$ denotes a matrix/tensor of ones of the appropriate size. We elaborate on our choices for the hyperparameters in the following paragraphs. For all other hyperparameters not explicitly specified below (such as the values of $\alpha_i$), we follow the default settings provided in Stable Diffusion 2's public  \href{https://github.com/Stability-AI/stablediffusion}{repository}\footnote{https://github.com/Stability-AI/stablediffusion}.

\paragraph{Adapting DDIM schedule}
We use DDIM \cite{song2021denoising} as the basis for configuring our sampler. We use the accelerated denoising process with $50$ time steps, uniformly spaced. We truncate the time-step range to $(300,1000)$ to prevent the network from focusing too much on artifacts introduced when rendering the latent texture map into latent images. At the last denoising step $i=1$, we perform sequential aggregation at the setting of $t_{i-1}=300$, but additionally compute $\rvx_0$ predictions $\hat{\rvx}_{0, n} = \frac{\rvx_{i,n} - \sqrt{1-\alpha_{i}} \epsilon_\theta^{t_i}(\rvx_{i, n})}{\sqrt{\alpha_i}}$ as final outputs. Following DDIM, we parameterize the noise scale of the DDIM process as $\sigma_i = \eta \sqrt{(1-\alpha_{i-1})/(1-\alpha_i)} \sqrt{1 - \alpha_i / \alpha_{i-1}}$. To maximize the consistency of updates produced in each viewpoint, we further introduce a temperature parameter $0 \leq \tau \leq 1$ which scales the noise term. Choosing $\tau < 1$ reduces the variance of the posterior $p(\rvx_{i-1} | \rvx_{i})$ without effecting its expectation. In the results presented in the manuscript, we use $\eta=1$, $\tau = 0.5$ in the coarse stage, and $\eta=1$, $\tau=0$ in the high-resolution refinement stage, which we find to be the most robust configuration.

\paragraph{Classifier-free guidance}
We use classifier-free guidance to control the alignment of texture to both depth and text. Specifically, we apply classifier-free guidance to both depth and text according to this formula: ${\epsilon'}_\theta^{t_i}(\rvx_{i, n};  d_n, \text{text}) = (1 - w_{joint}) \epsilon_\theta^{t_i}(\rvx_{i, n}) + w_{joint} \epsilon_\theta^{t_i}(\rvx_{i, n}; d_n, \text{text})$. We set $w_{joint} = 5$, and use ${\epsilon'}_\theta^{t_i}(\rvx_{i, n})$ in place of ${\epsilon}_\theta^{t_i}(\rvx_{i, n})$ in all experiments. We note that this formula is different from that used in SD2/w-depth, which only applies classifier-free guidance to the text prompt by including depth conditioning in both terms on the RHS of the equation.

For human heads and bodies, we find that stronger text guidance is helpful for stylization. Thus, we add a text-condition only term as follows: ${\epsilon'}_\theta^{t_i}(\rvx_{i, n};  d_n, \text{text}) = (1 - w_{joint} - w_{text}) \epsilon_\theta^{t_i}(\rvx_{i, n}) + w_{joint} \epsilon_\theta^{t_i}(\rvx_{i, n}; d_n, \text{text}) + w_{text} \epsilon_\theta^{t_i}(\rvx_{i, n}; \text{text})$. We set $w_{text}=3$ for these prompts.

\paragraph{Additional geometry processing} 
We align objects with meaningful ``front" to face the $+x$ direction, and ensure all objects are placed with $+y$ as ``up". Following \cite{poole2022dreamfusion}, we augment prompts with ``\{prompt\}, front/side/rear/top-view" based on the location of the camera to the nearest exact direction; ``top-view" is used when the elevation of the camera is above $60^{\circ}$. Perspective cameras are placed facing the origin at a fixed distance of $1.5$ from the origin, and adjust the FOV to fit the object within the image. For most objects, we find that a set of nine cameras - all looking at the origin, eight spaced uniformly surrounding the object (azimuth from $0^\circ$ to $315^\circ$ spaced $45^\circ$ apart, at an elevation of $30^\circ$), and one camera looking down the $-y$ direction - to work reasonable well for objects with reasonable aspect ratios and few occlusions. 

In the first round of SIMS sampling, we apply $10^\circ$ random jitters to the elevation and azimuth of each camera, and re-sample each camera for a total of $18$ cameras to ensure surface coverage. In the second round, we do not apply jittering and use the fixed set of nine cameras. For human characters, the default set of nine cameras does not adequately cover the entire surface due to occlusions. We instead use 3 sets of 8 cameras: each set is placed radially looking at the $y$ axis (azimuth from $0^\circ$ to $315^\circ$ spaced $45^\circ$ apart), and a different offset is applied to the cameras' $y$ position depending on the set ($0.3, 0.0, -0.3$ respectively). This forms a cylinder of cameras looking at the $y$ axis, and adequately covers all surfaces on the human character geometry.

\begin{algorithm*}[h]
\caption{Sequential Interlaced Multiview Sampler}\label{alg:sims}
\begin{algorithmic}
\STATE Input: mesh $\mathcal{M}$, cameras $\{C_1, \dots, C_N \}$
\STATE Parameters: Denoising time schedule $\{t_i\}_{i=T}^0$, DDIM noise schedule $\{\sigma_i\}_{i=T}^0$, DDIM noise scale $\eta$, temperature $\tau$, function for camera jittering $maybe\_apply\_jitter$
\STATE $\rvz_T \sim \mathcal{N}(\mathbf{0}, \boldsymbol{I})$
\FOR{$i \in \{T \dots 1\}$}
    {\color{Orchid}
    \STATE Init mask $M_i = 0$ of shape $(N, H, W)$
    \STATE Init quality buffer $Q_i = - \infty$ of shape $(N, H, W)$
    \STATE $\rvz_{i-1, 0} = \rvz_i$
    \STATE Apply camera jitter $\{C_{i,1}, \dots, C_{i,N} \} = maybe\_apply\_jitter(\{C_1, \dots, C_N \})$
    \STATE Sample forward noise $\epsilon_i$}
    \FOR{$n \in \{1 \dots N\}$}
        {\color{YellowGreen}
        \STATE Compute forward diffusion term $\rvz_{i, n} = M_{i} \odot \left(\sqrt{\frac{\alpha_{i-1}}{\alpha_i}}  \rvz_{i-1, n-1} +\sqrt{1 - \frac{\alpha_t}{\alpha_{t-1}}} \epsilon_i \right) + \left( \mathbf{1} - M_{i}\right) \odot \rvz_i$ 
        \STATE Render latent image and compute screen space derivatives $\rvx_{i, n}', (\frac{\partial \rvu}{\partial \rvp}, \frac{\partial \rvv}{\partial \rvp}, \frac{\partial \rvu}{\partial \rvq}, \frac{\partial \rvv}{\partial \rvq})= \mathcal{R}(\rvz_{i,n}; C_{i,n})$
        \STATE $J_{i,n} = \left| \frac{\partial \rvu}{\partial \rvp} \cdot \frac{\partial \rvv}{\partial \rvq} - \frac{\partial \rvu}{\partial \rvq} \cdot  \frac{\partial \rvv}{\partial \rvp} \right|$}
        {\color{CornflowerBlue}
        \STATE Sample $\varepsilon_{i,n} \sim \mathcal{N}(\mathrm{0}, \boldsymbol{I})$
        \STATE Perform denoising: $\rvx_{i-1, n} = \sqrt{\alpha_{i-1}} \left(\frac{\rvx_{i,n} - \sqrt{1-\alpha_{i}} \epsilon_\theta^{t_i}(\rvx_{i, n})}{\sqrt{\alpha_i}} \right) + \sqrt{1 - \alpha_{i-1} - \sigma_{i}^2} \cdot \epsilon_\theta^{t_i}(\rvx_{i, n}) + \tau \cdot \sigma_{i} \cdot \varepsilon_{i,n} $
        \IF{$i = 1$}
            \STATE $\rvx_0$ prediction: $\hat{\rvx}_{0, n} = \frac{\rvx_{i,n} - \sqrt{1-\alpha_{i}} \epsilon_\theta^{t_i}(\rvx_{i, n})}{\sqrt{\alpha_i}}$
        \ENDIF}
        {\color{BrickRed}
        \STATE $\rvz_{i-1, n}' = \mathcal{R}^{-1}(\rvx_{i-1, n}; C_{i,n})$
        \STATE $Q_{i,n} = \mathcal{R}^{-1}(- J_{i-1, n}; C_{i,n})$
        \STATE $M_{i,n} = \mathcal{R}^{-1}(\boldsymbol{I}(\rvx_{i-1,n}); C_{i,n})$
        
        \STATE Determine update area $U = M_{i,n}(u,v) > 0 \text{, and} \, Q_{i,n} > Q_i$
        \STATE Update pixels $\rvz_{i-1, n} = U \odot \rvz_{i-1, n}' + (1 - U) \odot \rvz_{i-1, n}$
        \STATE Update mask and quality buffer $M_i = \max{(M_i, M_{i,n})}$,  $Q_i = \max{(Q_i, Q_{i,n})}$ (max is applied element-wise)}
    \ENDFOR
    \STATE {$\rvz_{i-1} = \rvz_{i-1, N}$}
\ENDFOR
\RETURN $\{\hat{\rvx}_{0, n}\}_{n=1}^N$
\end{algorithmic}
\end{algorithm*}

\section{Additional Results}

\subsection{Qualitative Results}
\begin{figure}[h!]

    \centering
    \vspace{-0.3cm}
    \setlength{\tabcolsep}{1pt}
    {\small 
    \renewcommand{\arraystretch}{0.8}
    \begin{tabular}{c c c}
        Stable-DreamFusion & Latent-Painter & \ourmodel \\
        \includegraphics[width=0.3\linewidth]{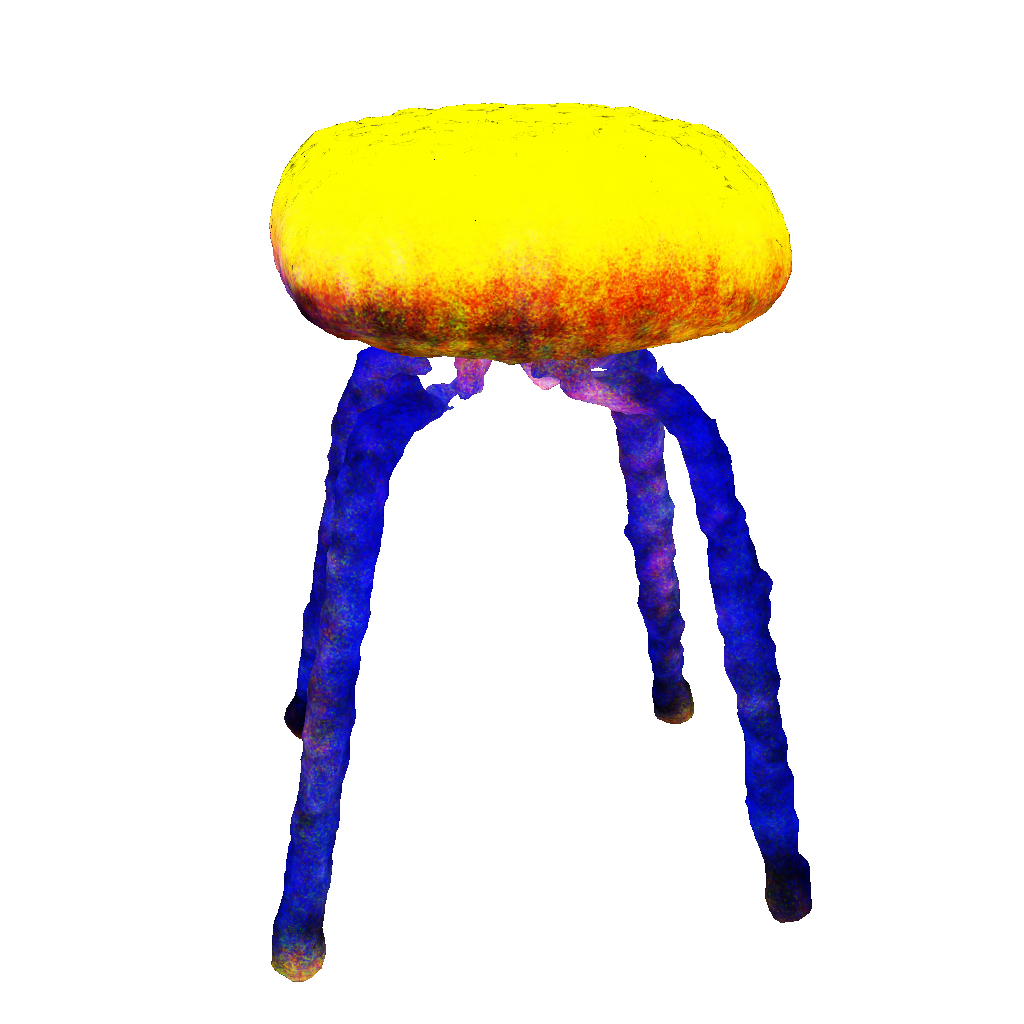} &
        \includegraphics[width=0.3\linewidth]{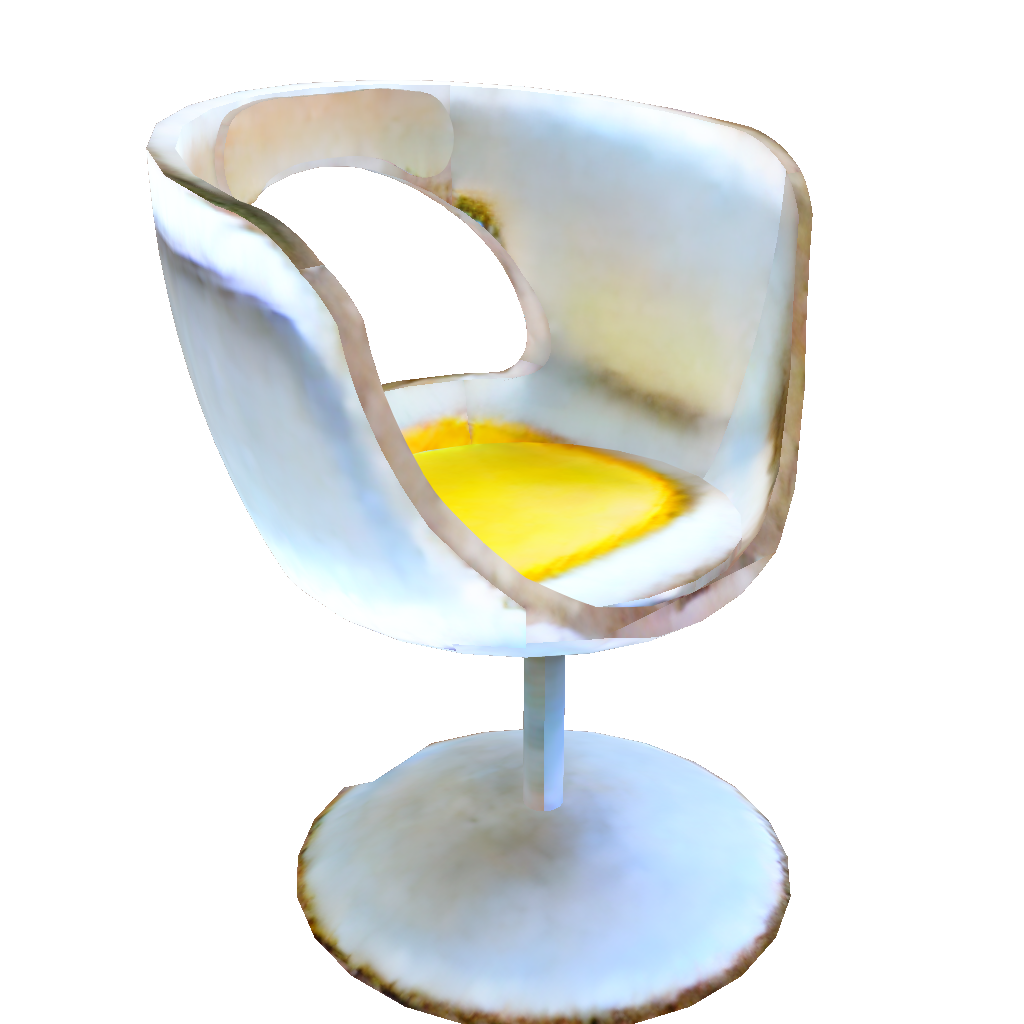} &
        \includegraphics[width=0.3\linewidth]{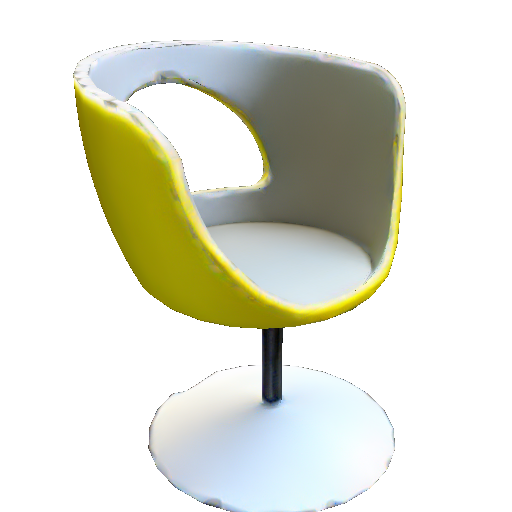} \\
        \multicolumn{3}{c}{``yellow plastic stool with white seat"} \vspace{3pt}\\
        \includegraphics[width=0.3\linewidth]{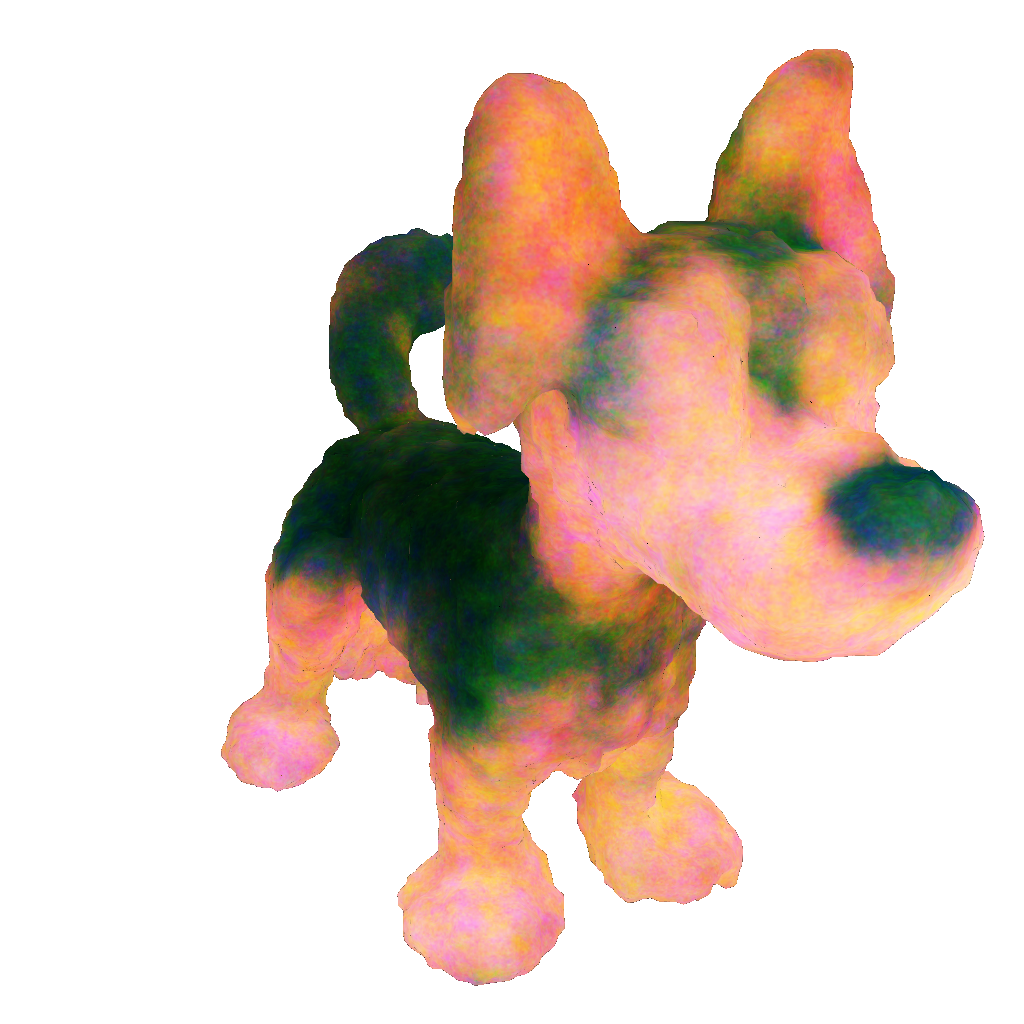} &
        \includegraphics[width=0.3\linewidth]{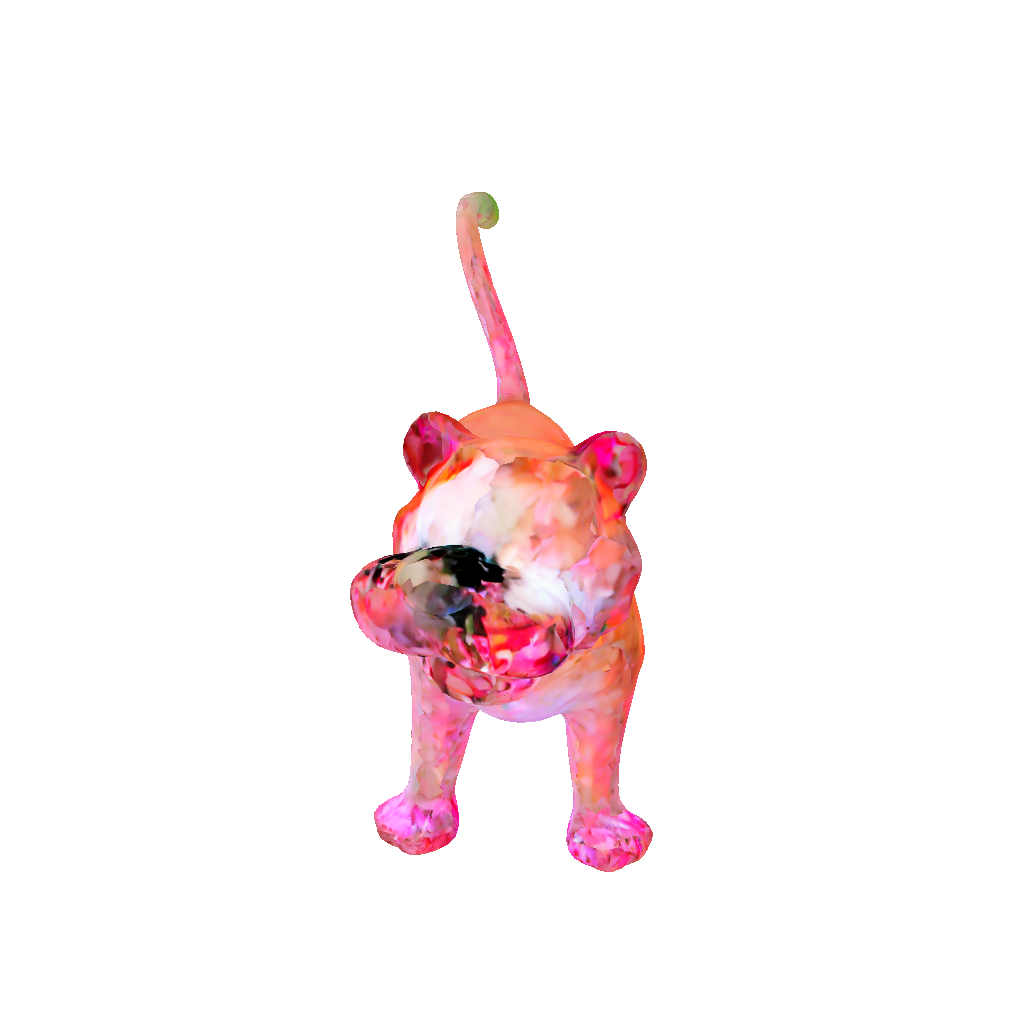} &
        \includegraphics[width=0.3\linewidth]{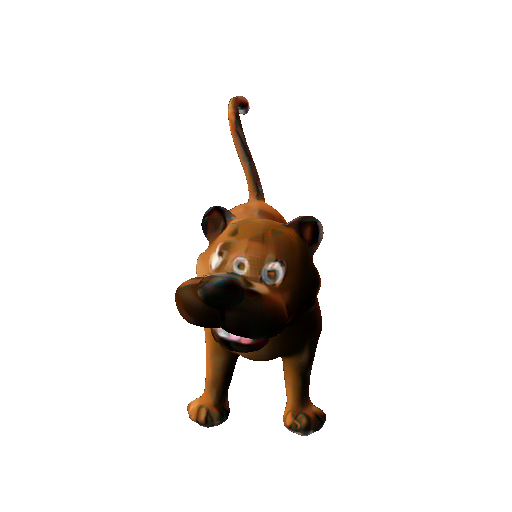} \\
        \multicolumn{3}{c}{``cartoon dog"} \vspace{3pt}\\
        \includegraphics[width=0.3\linewidth]{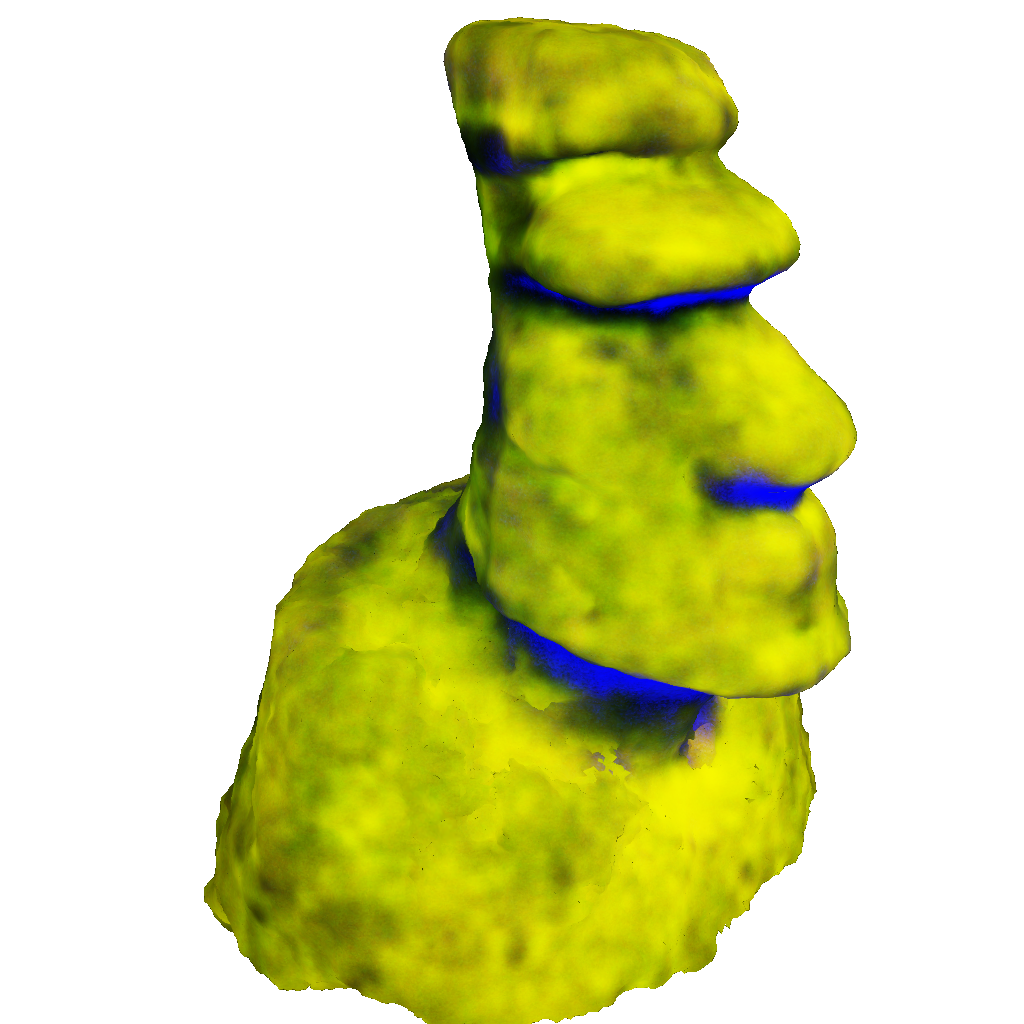} &
        \includegraphics[width=0.3\linewidth]{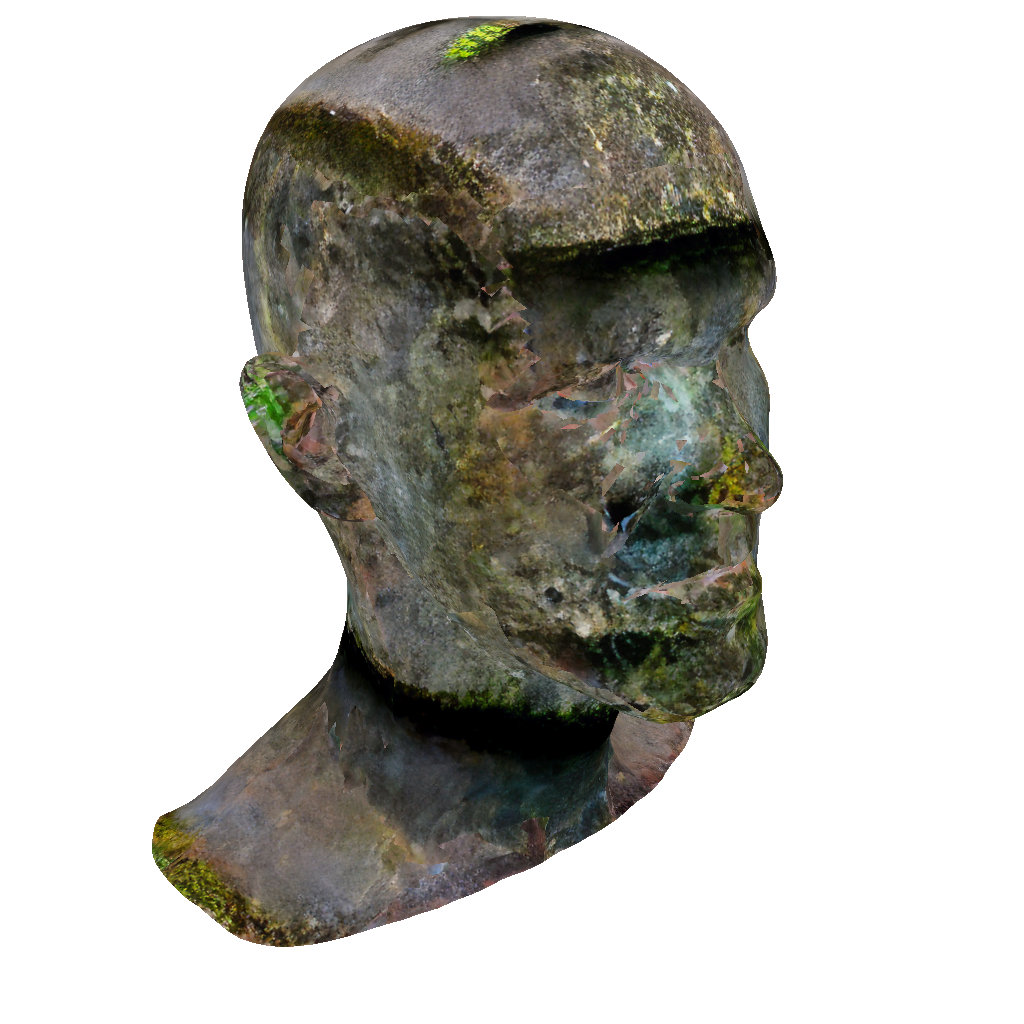} &
        \includegraphics[width=0.3\linewidth]{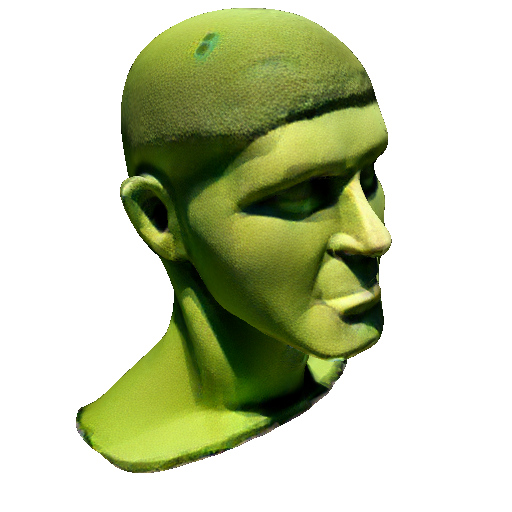} \\
        \multicolumn{3}{c}{``moai stone statue with green moss on top"} \vspace{3pt}\\
        \includegraphics[width=0.3\linewidth]{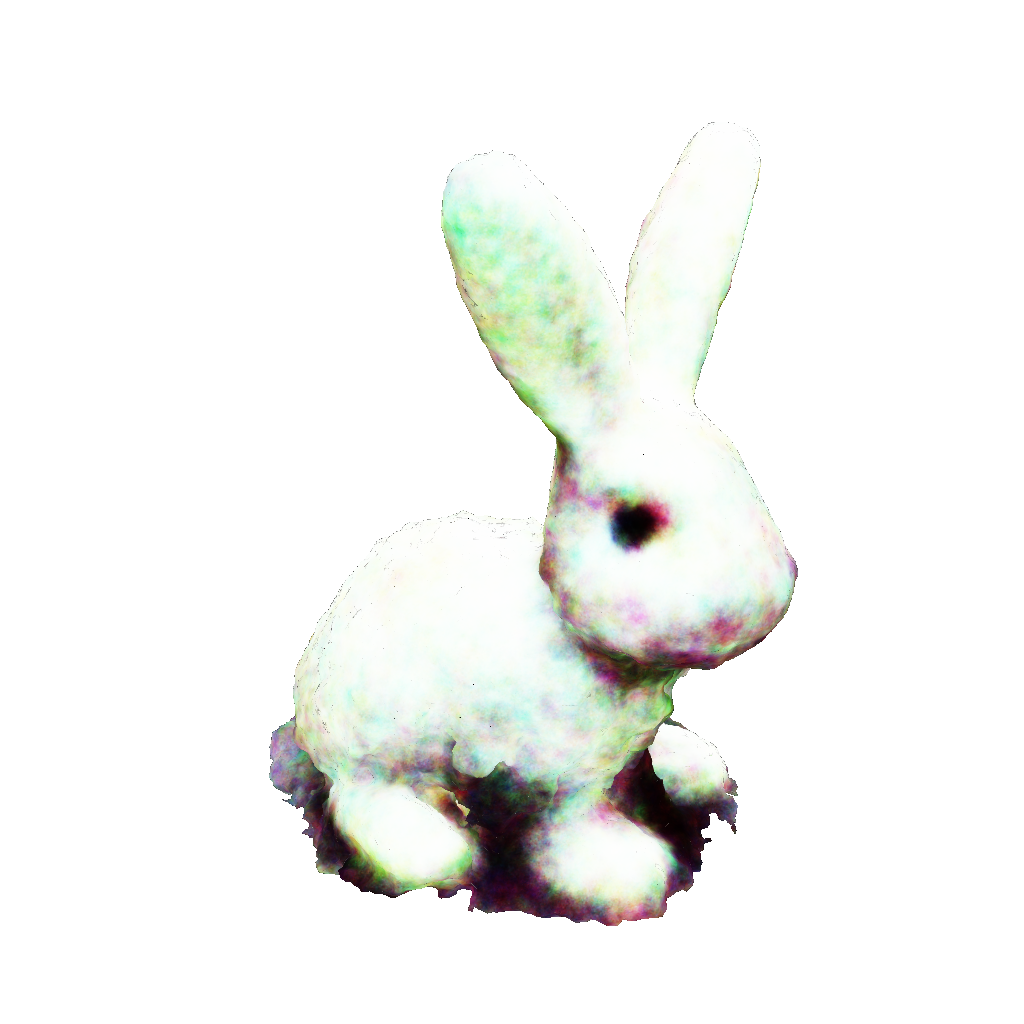} &
        \includegraphics[width=0.3\linewidth]{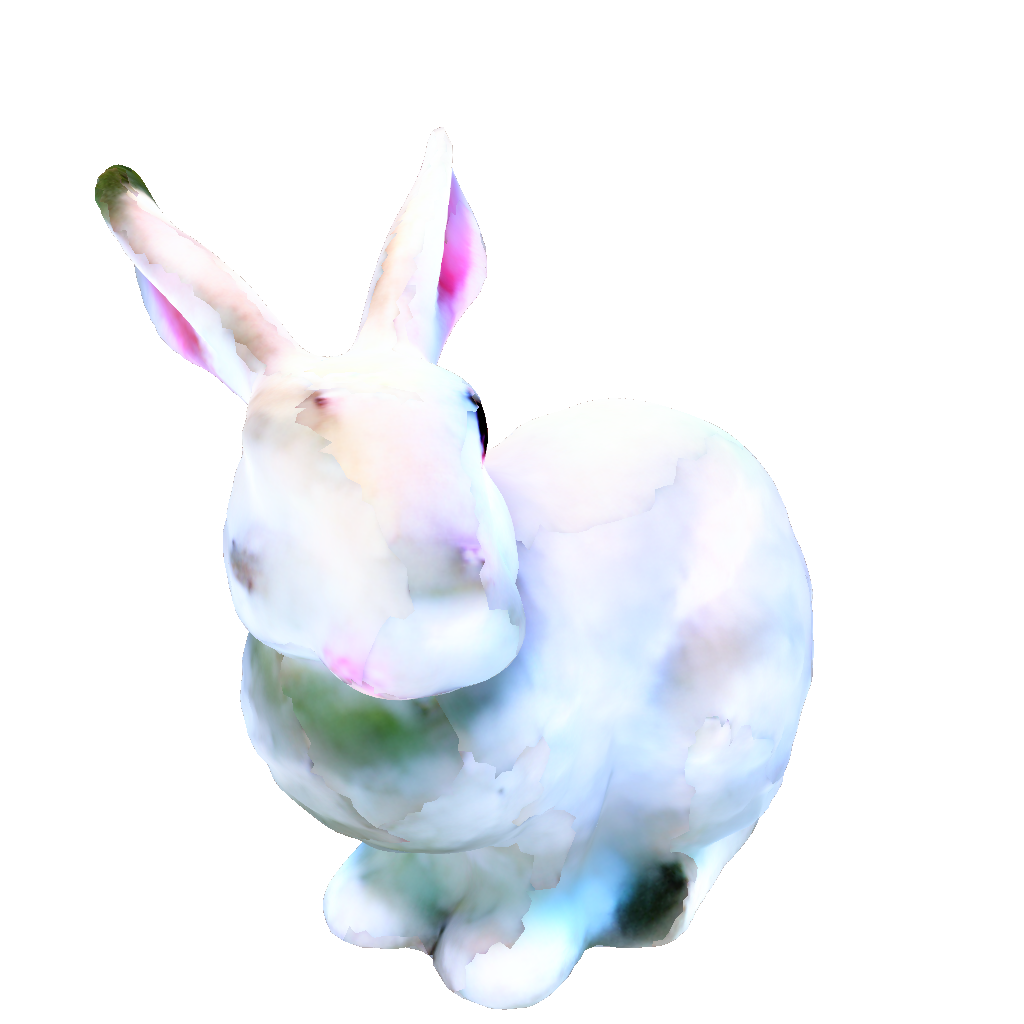} &
        \includegraphics[width=0.3\linewidth]{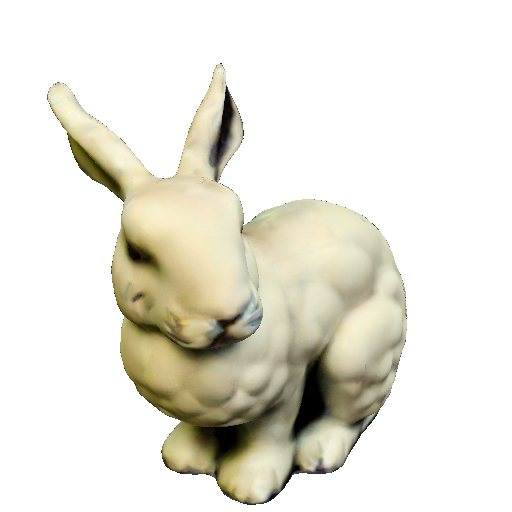} \\
        \multicolumn{3}{c}{``white bunny"} \vspace{3pt}\\
        \includegraphics[width=0.3\linewidth]{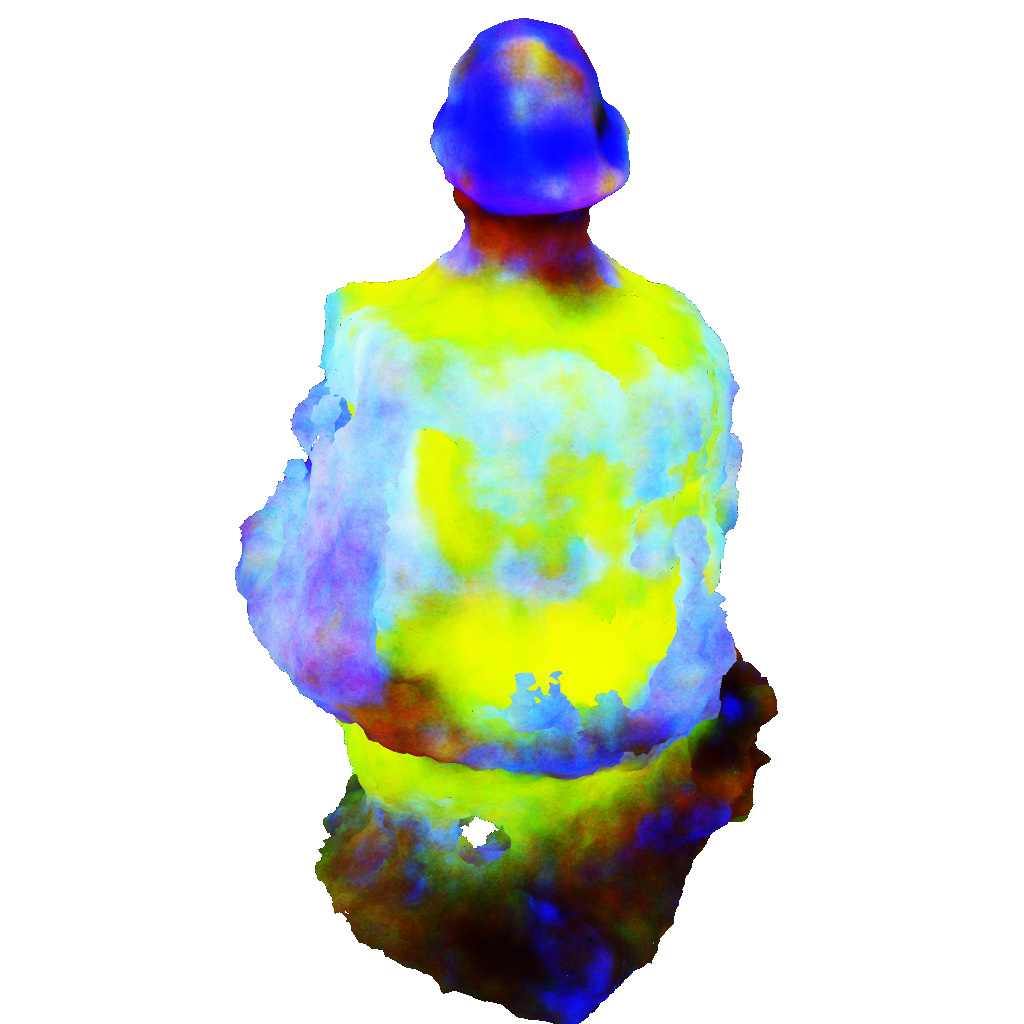} &
        \includegraphics[width=0.3\linewidth]{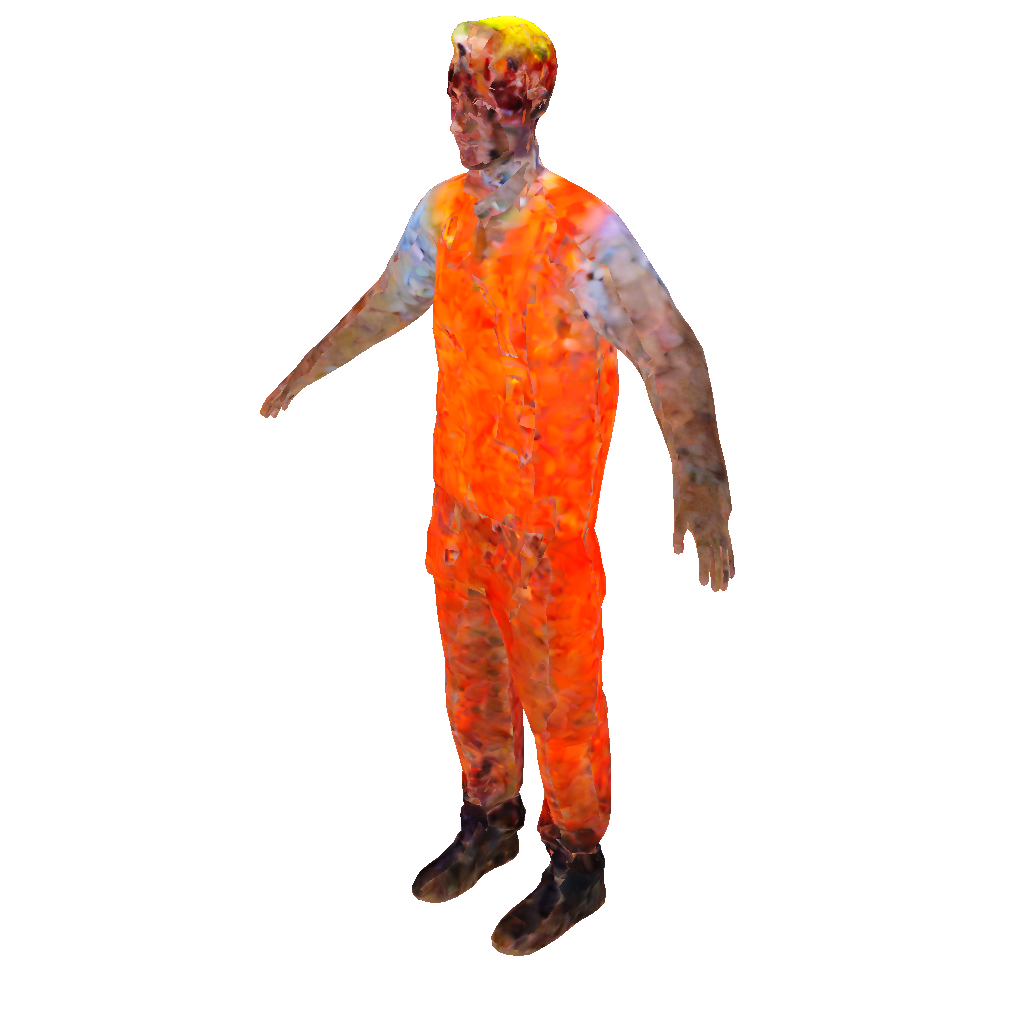} &
        \includegraphics[width=0.3\linewidth]{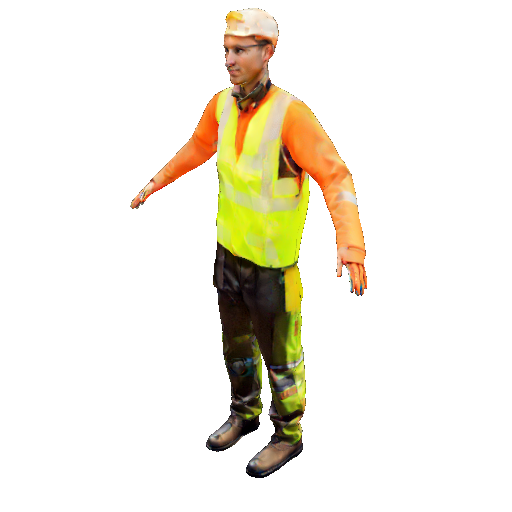} \\
        \multicolumn{3}{c}{``railroad worker wearing high-vis vest"} \vspace{3pt}\\
         \includegraphics[width=0.3\linewidth]{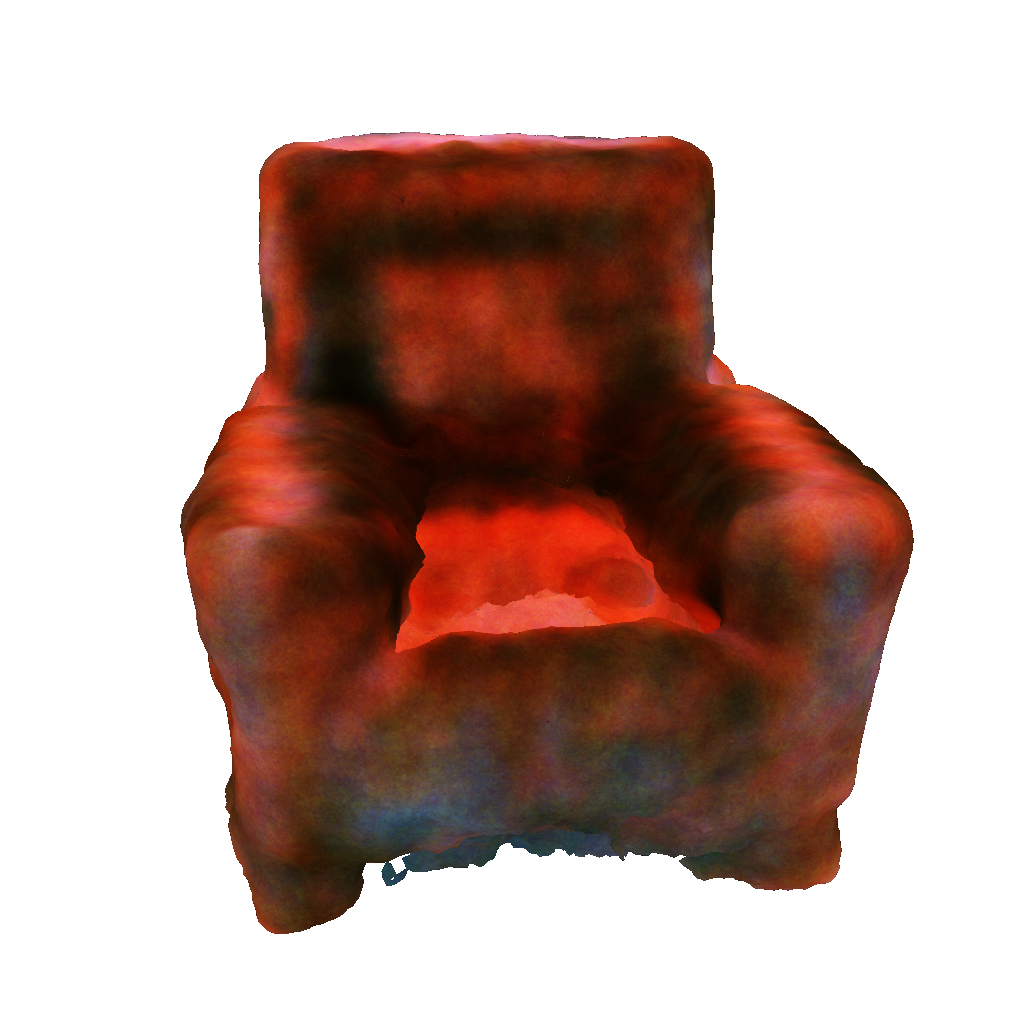} &
        \includegraphics[width=0.3\linewidth]{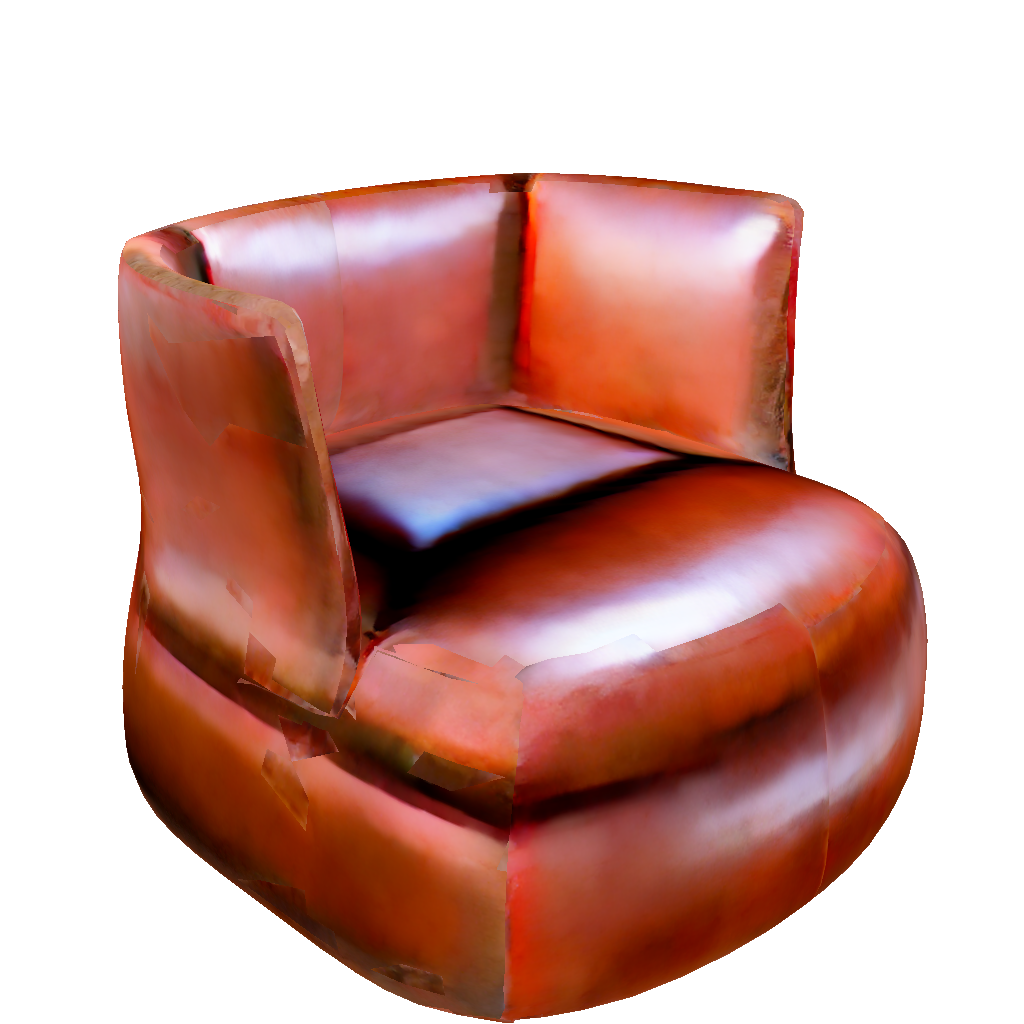} &
        \includegraphics[width=0.3\linewidth]{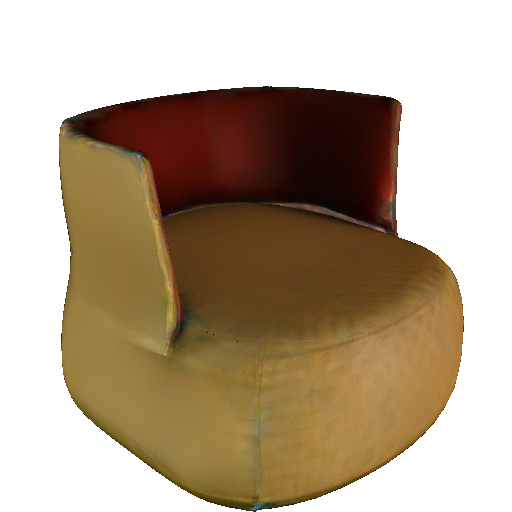} \\
        \multicolumn{3}{c}{``leather lounge chair"}\\
        
    \end{tabular}}
    
    \caption{Visual comparison of texture generated by Stable DreamFusion (left) \cite{stable-dreamfusion}, Latent-Painter (middle) \cite{metzer2022latent}, and our \ourmodel(right). Prompts are cherry picked for those where Stable DreamFusion successfully converged to a reasonable geometry. 
    \vspace{-4mm}
    }
    \label{fig:new_baseline}
\end{figure}

\begin{figure}[t]

    \centering
    \vspace{-0.15cm}
    \setlength{\tabcolsep}{1pt}
    {\small 
    \renewcommand{\arraystretch}{0.8}
    \begin{tabular}{c c c c}
              \includegraphics[width=0.24\linewidth]{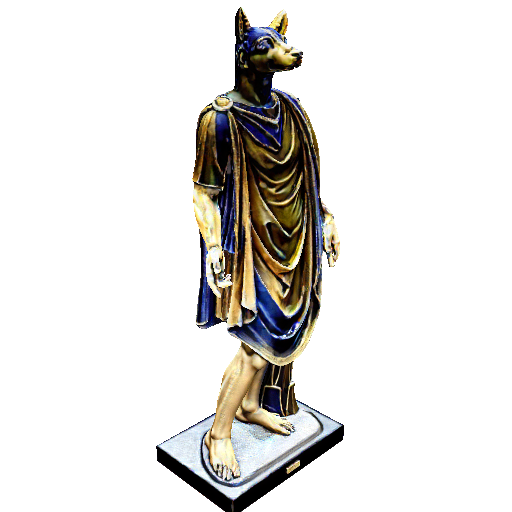} &
        \includegraphics[width=0.24\linewidth]{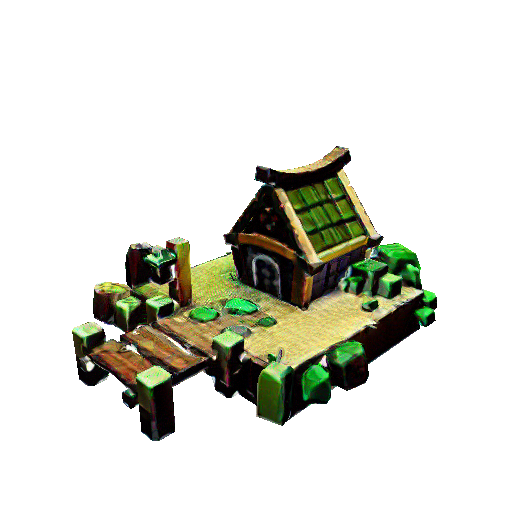} &
        \includegraphics[width=0.24\linewidth]{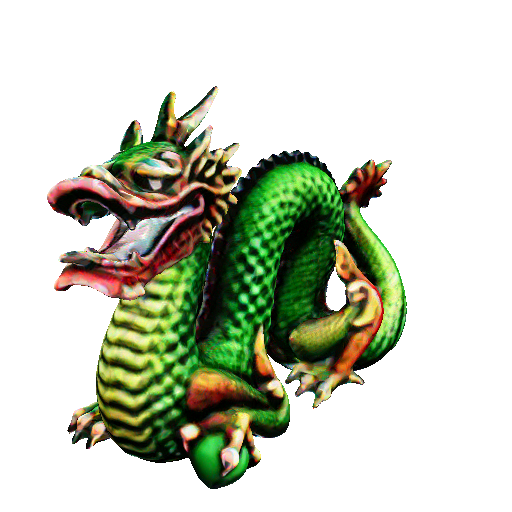} &
        \includegraphics[width=0.24\linewidth]{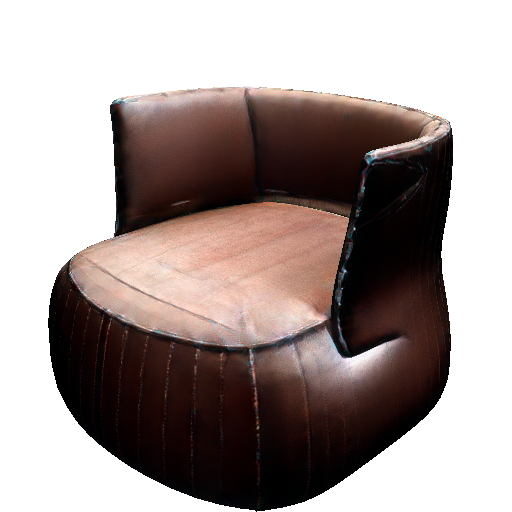} \\
                \includegraphics[width=0.24\linewidth]{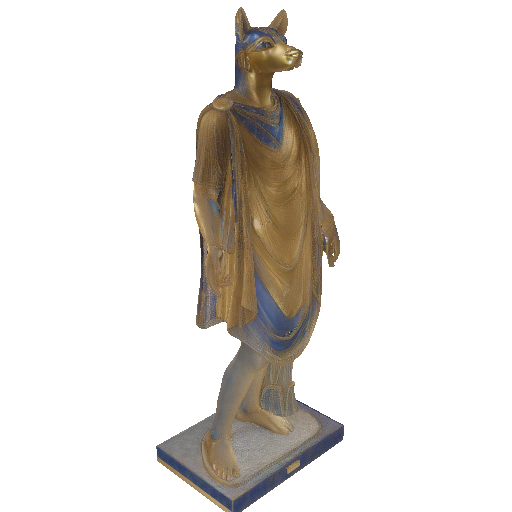} &
         \includegraphics[width=0.24\linewidth]{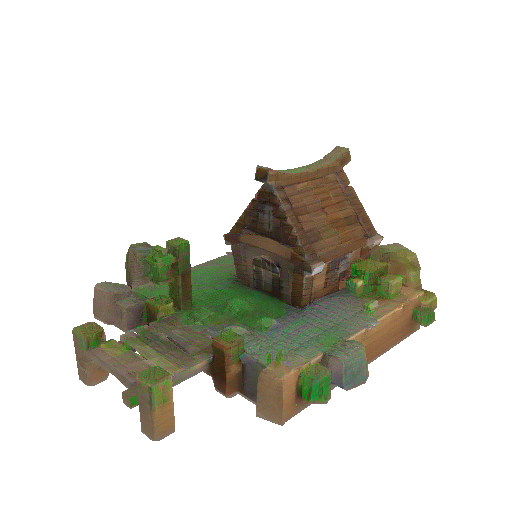} &
        \includegraphics[width=0.24\linewidth]{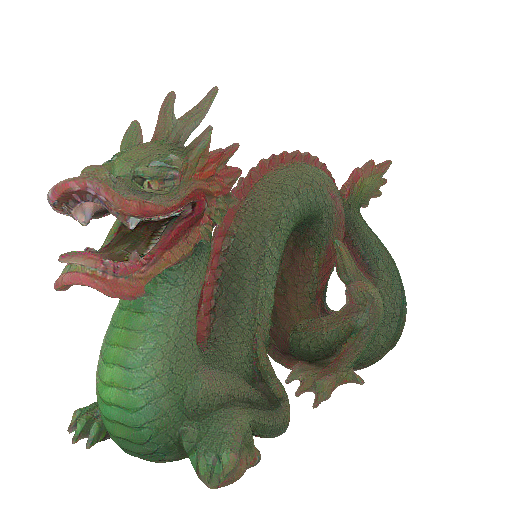} &

            \includegraphics[width=0.24\linewidth]{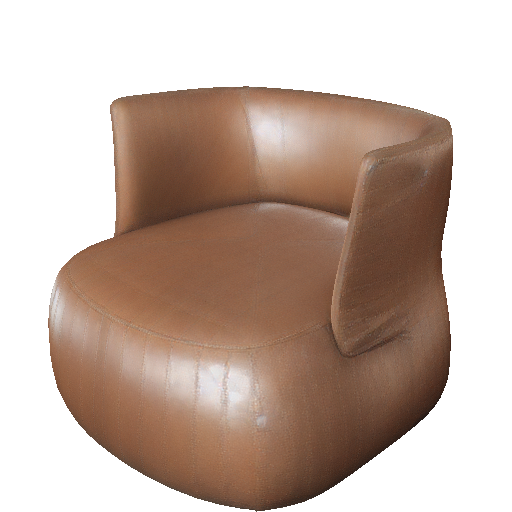}

        
    \end{tabular}}
    
    \caption{\small Top: TexFusion + ControlNet in ``guess mode"; bottom: TexFusion + ControlNet in ``normal mode".
    \vspace{-3mm}
    }
    \label{fig:even_more_results_cnet}
\end{figure}

\begin{figure*}[t!]

    \centering
    \vspace{-0.15cm}
    \setlength{\tabcolsep}{1pt}
    {\small 
    \renewcommand{\arraystretch}{0.8}
    \begin{tabular}{c c c c c c}
    
        \includegraphics[width=0.15\linewidth]{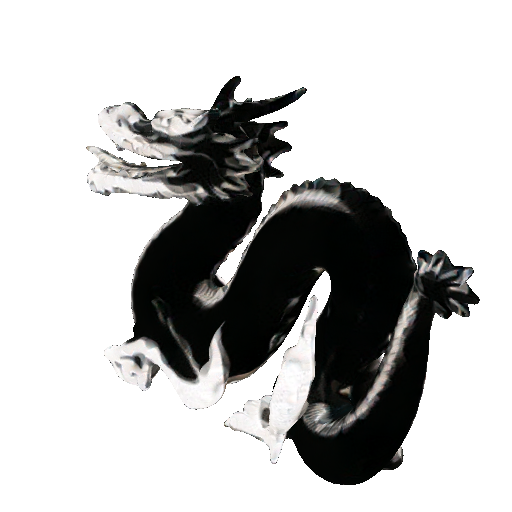} &
        \includegraphics[width=0.15\linewidth]{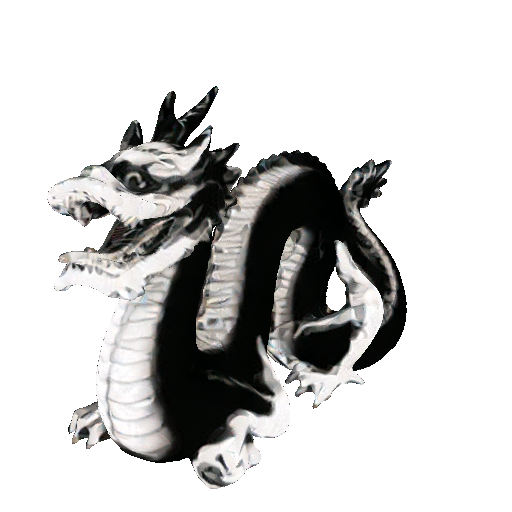} &
        \includegraphics[width=0.15\linewidth]{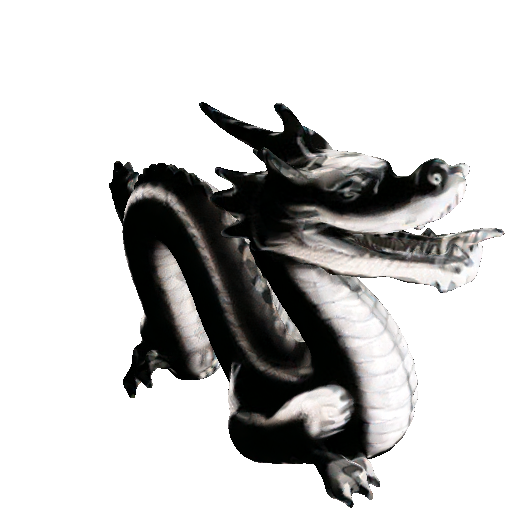} &
        \includegraphics[width=0.15\linewidth]{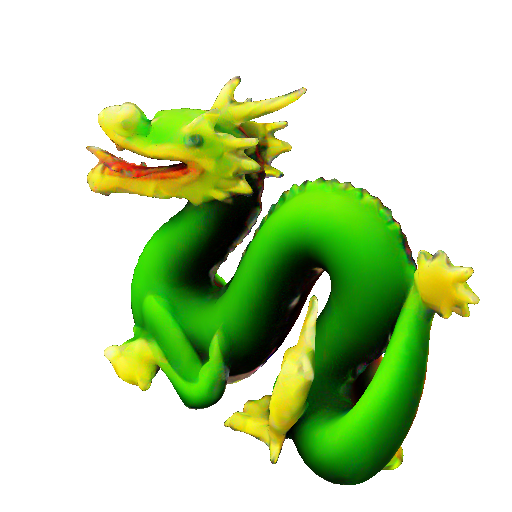} &
        \includegraphics[width=0.15\linewidth]{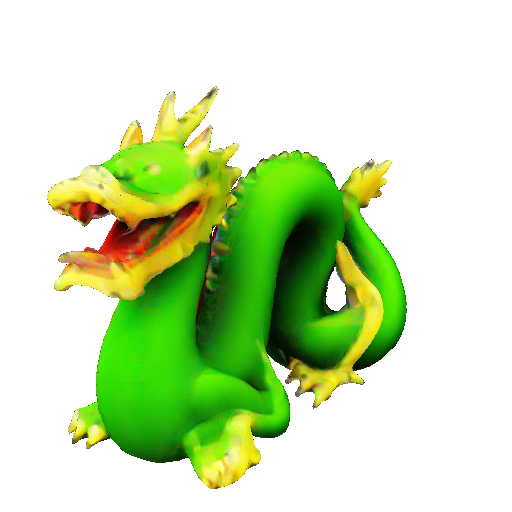} &
        \includegraphics[width=0.15\linewidth]{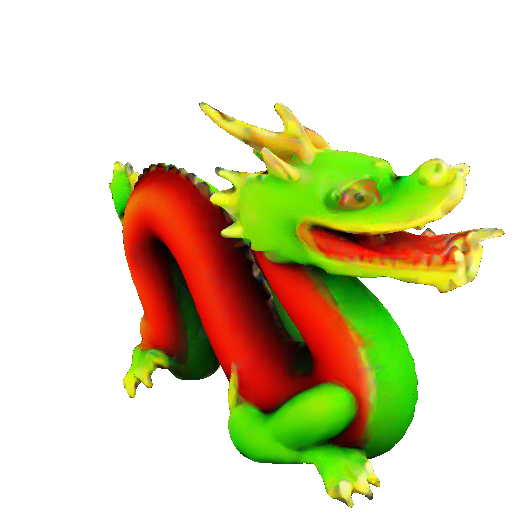} \\
        \multicolumn{3}{c}{``black and white dragon in chinese ink art style"} & \multicolumn{3}{c}{``cartoon dragon, red and green"}\\
        
        \includegraphics[width=0.15\linewidth]{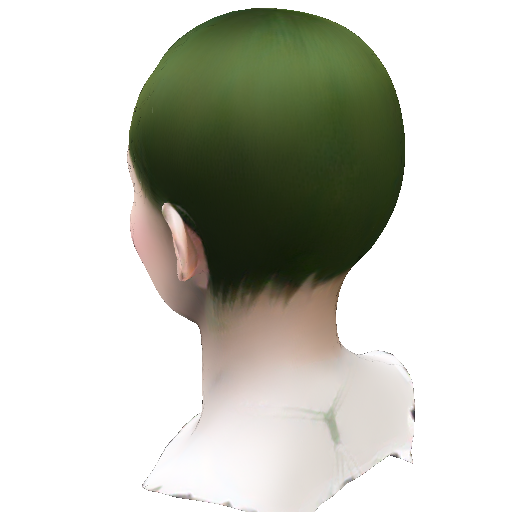} &
        \includegraphics[width=0.15\linewidth]{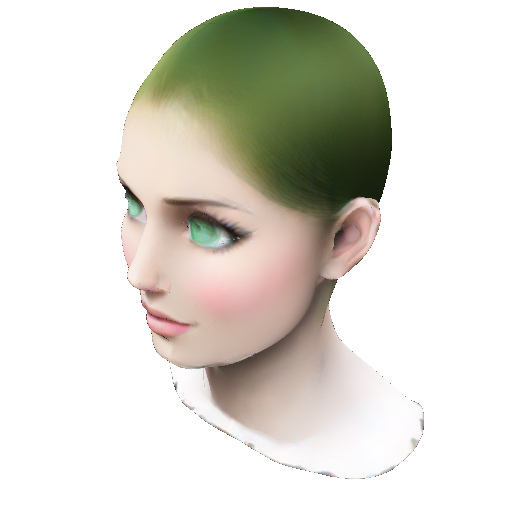} &
        \includegraphics[width=0.15\linewidth]{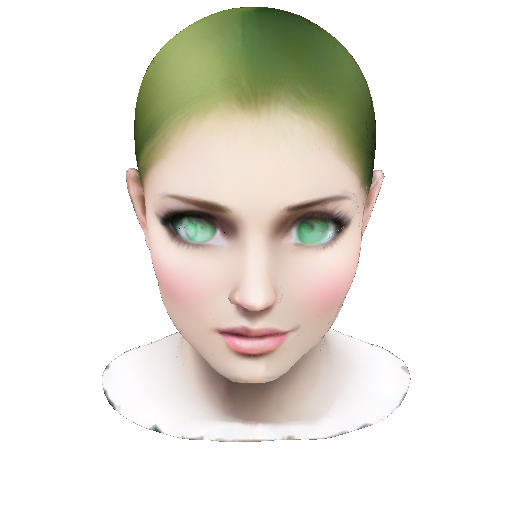} &
        \includegraphics[width=0.15\linewidth]{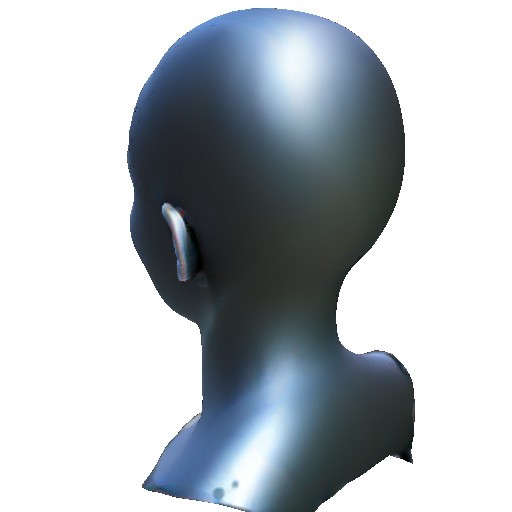} &
        \includegraphics[width=0.15\linewidth]{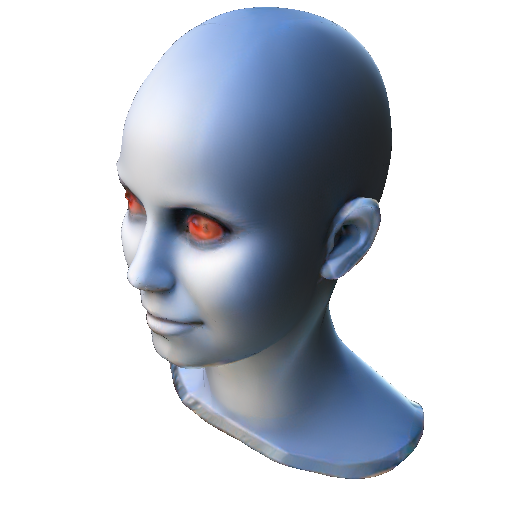} &
        \includegraphics[width=0.15\linewidth]{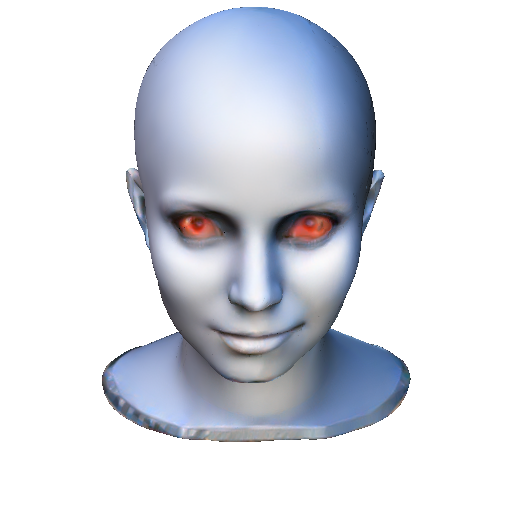} \\
        \multicolumn{3}{c}{``blonde girl with green eyes, hair in a tied } & \multicolumn{3}{c}{``Portrait of a humanoid robot, futuristic, }\\
        \multicolumn{3}{c}{bun, anime illustration, portrait"} & \multicolumn{3}{c}{science fiction"} \\
         \includegraphics[width=0.15\linewidth]{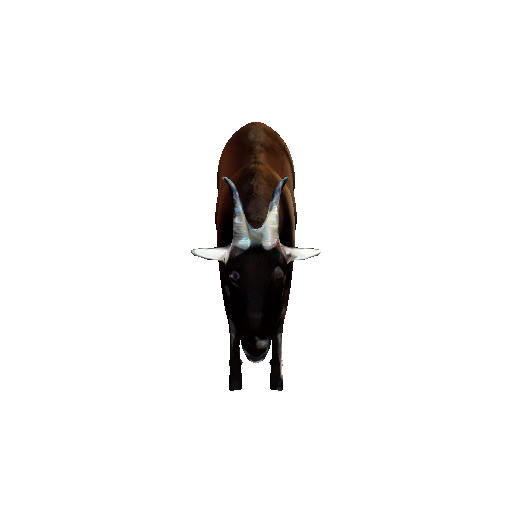} &
         \includegraphics[width=0.15\linewidth]{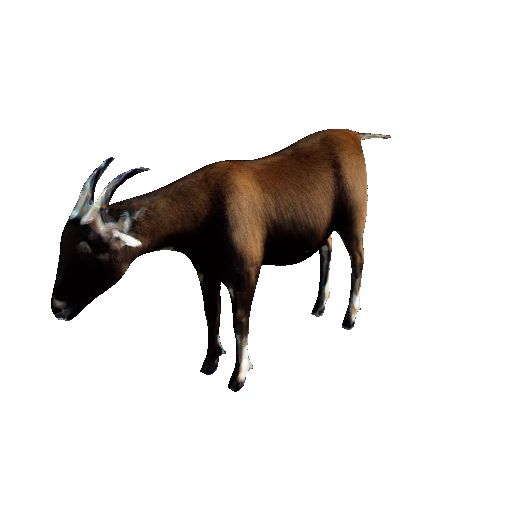} &
        \includegraphics[width=0.15\linewidth]{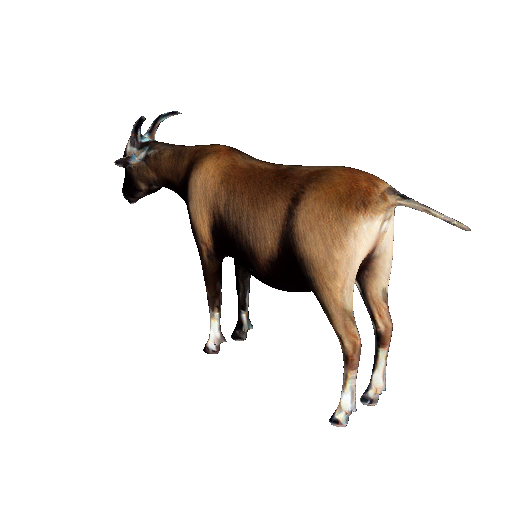} &
        \includegraphics[width=0.15\linewidth]{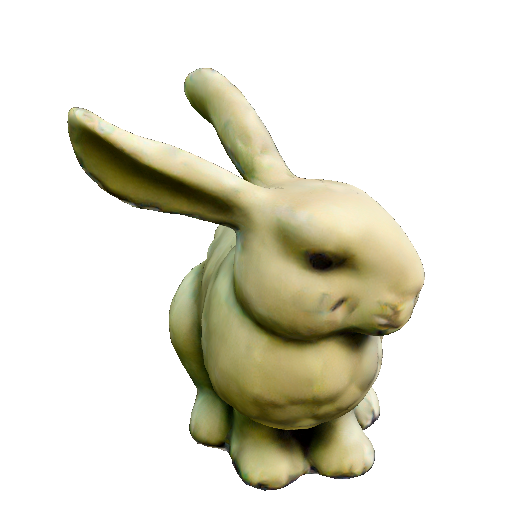} &
        \includegraphics[width=0.15\linewidth]{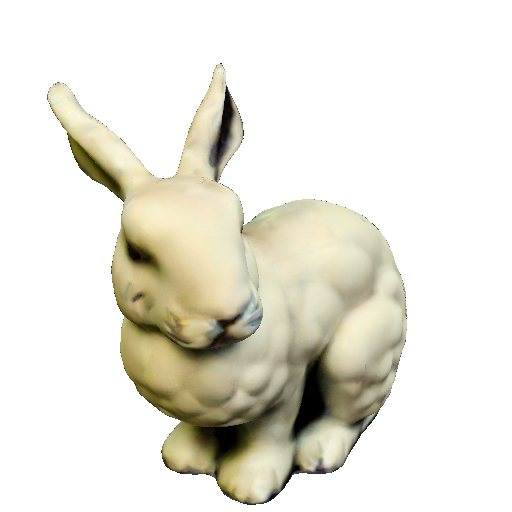} &
        \includegraphics[width=0.15\linewidth]{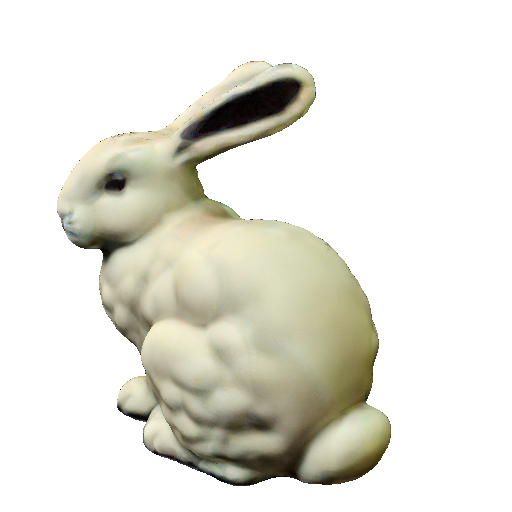} \\
        \multicolumn{3}{c}{``brown mountain goat"} & \multicolumn{3}{c}{``white bunny"}\\
         \includegraphics[width=0.15\linewidth]{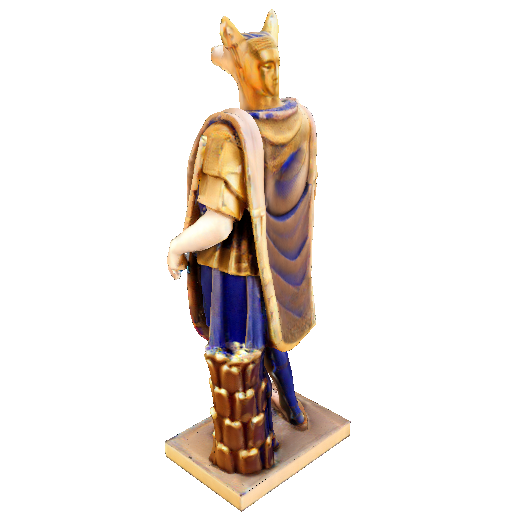} &
         \includegraphics[width=0.15\linewidth]{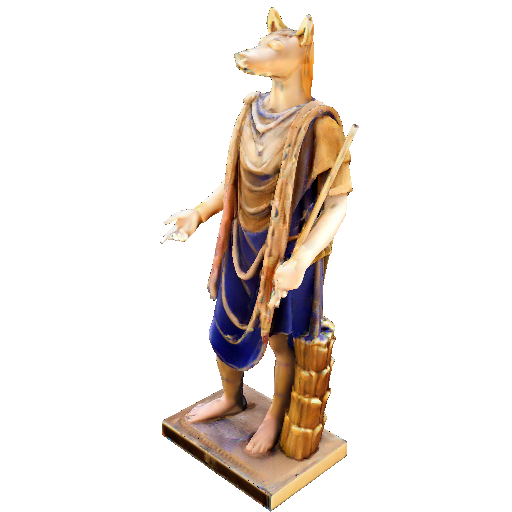} &
        \includegraphics[width=0.15\linewidth]{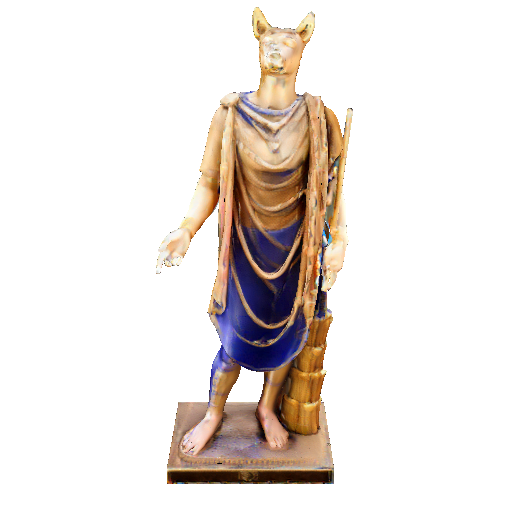} &
        \includegraphics[width=0.15\linewidth]{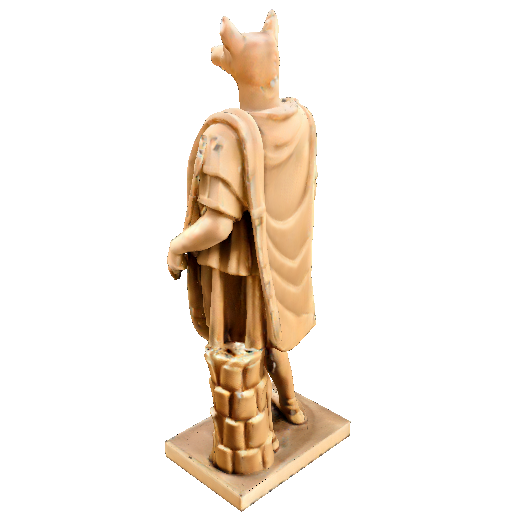} &
        \includegraphics[width=0.15\linewidth]{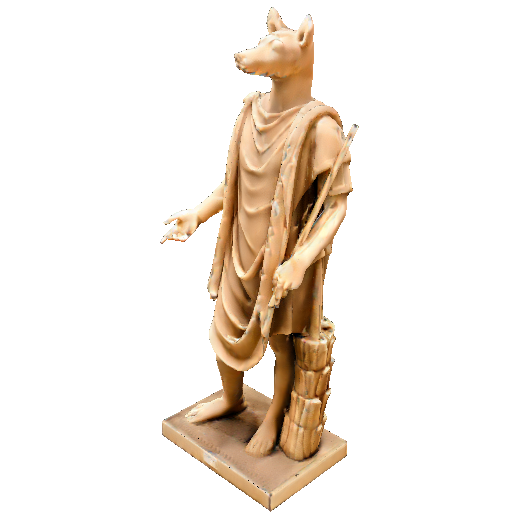} &
        \includegraphics[width=0.15\linewidth]{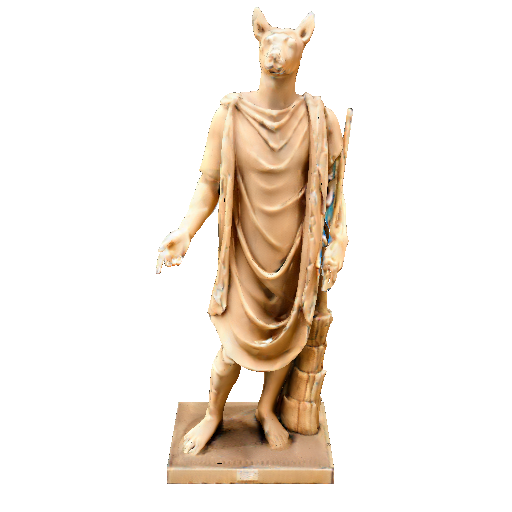} \\
               \multicolumn{3}{c}{``portrait of greek-egyptian deity hermanubis,  } & \multicolumn{3}{c}{\multirow{2}{*}{``sandstone statue of hermanubis"}}\\
        \multicolumn{3}{c}{lapis skin and gold clothing"} & \multicolumn{3}{c}{} \\
         \includegraphics[width=0.15\linewidth]{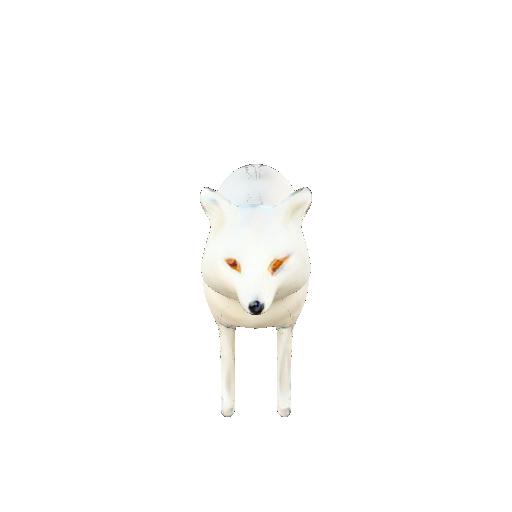} &
         \includegraphics[width=0.15\linewidth]{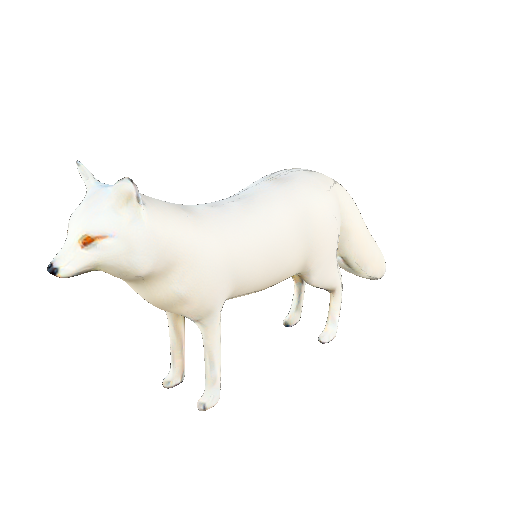} &
         \includegraphics[width=0.15\linewidth]{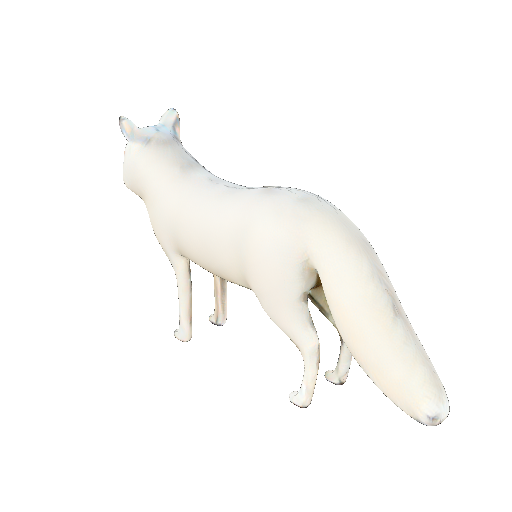} &
         
         \includegraphics[width=0.15\linewidth]{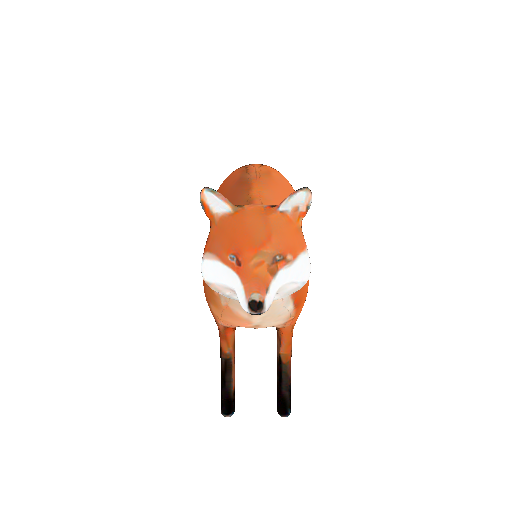} &
         \includegraphics[width=0.15\linewidth]{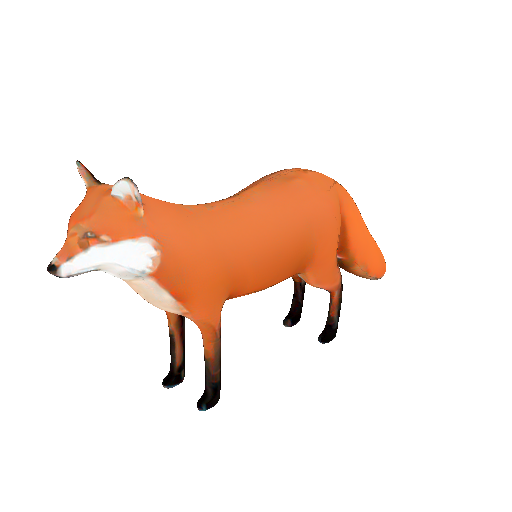} &
         \includegraphics[width=0.15\linewidth]{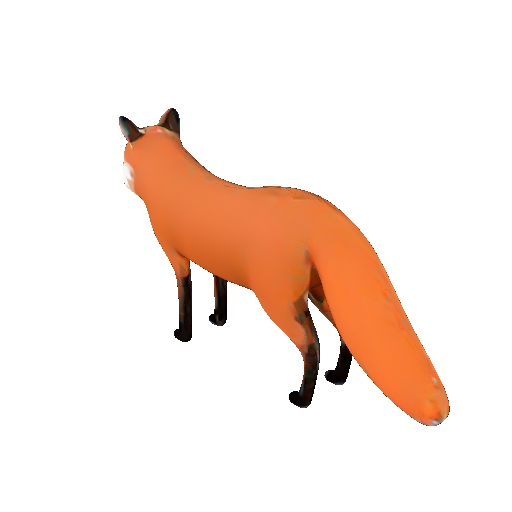} \\
        \multicolumn{3}{c}{``white fox"} & \multicolumn{3}{c}{``cartoon fox"}\\
        \includegraphics[width=0.15\linewidth]{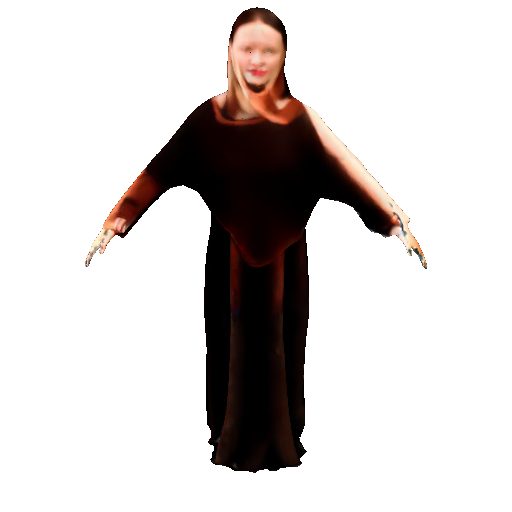} &
        \includegraphics[width=0.15\linewidth]{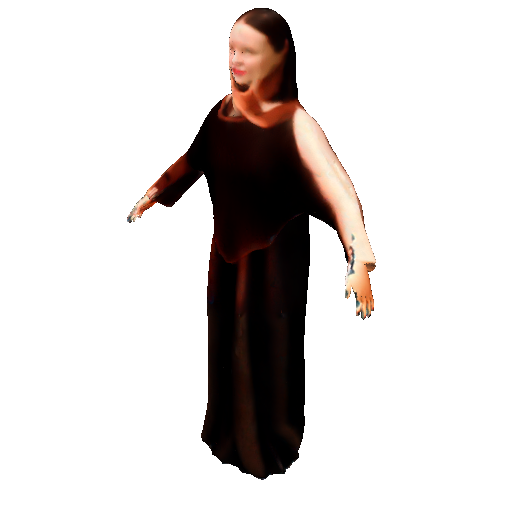} &
        \includegraphics[width=0.15\linewidth]{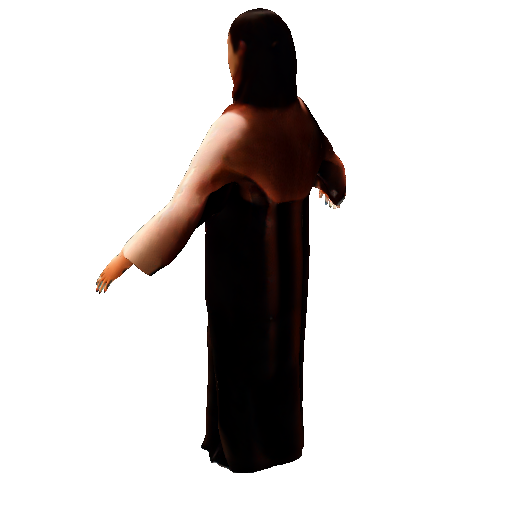} &
        \includegraphics[width=0.15\linewidth]{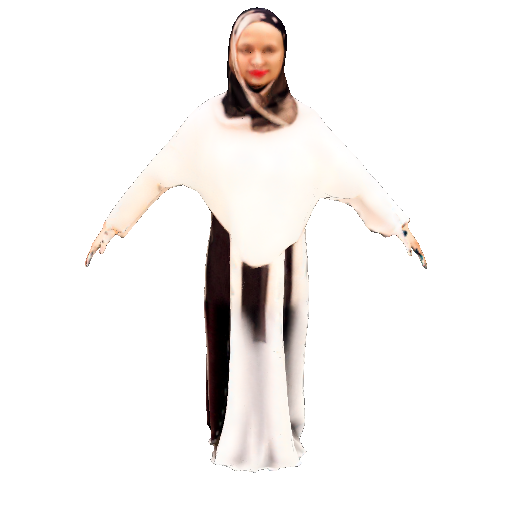} &
        \includegraphics[width=0.15\linewidth]{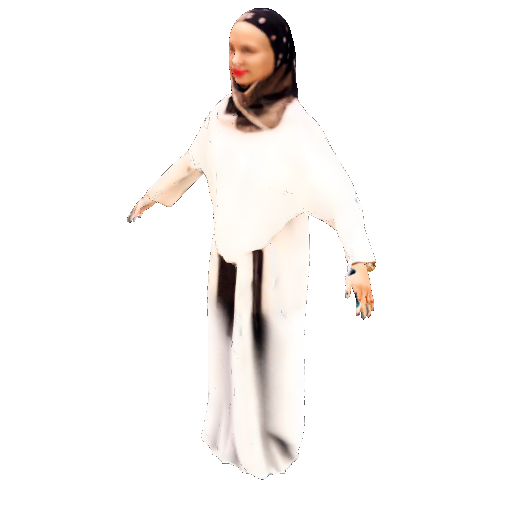} &
        \includegraphics[width=0.15\linewidth]{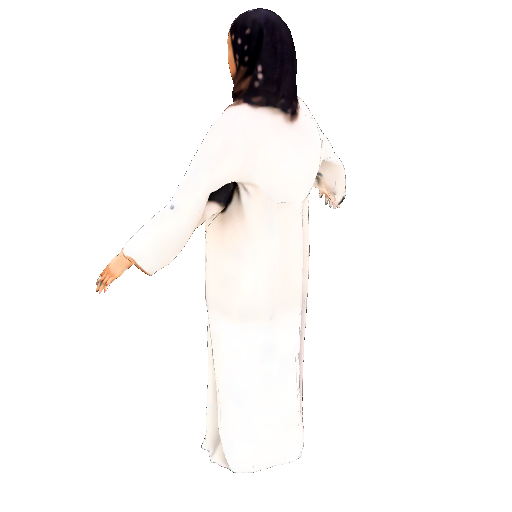} \\
        \multicolumn{3}{c}{``nunn in a black dress"} & \multicolumn{3}{c}{``nunn in a white dress, black headscarf"}\\
        \includegraphics[width=0.15\linewidth]{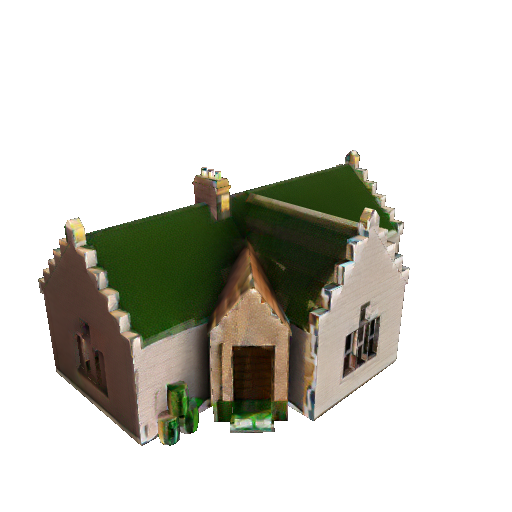} &
        \includegraphics[width=0.15\linewidth]{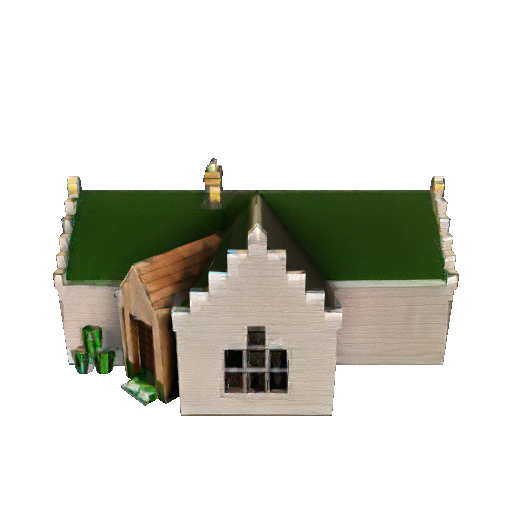} &
        \includegraphics[width=0.15\linewidth]{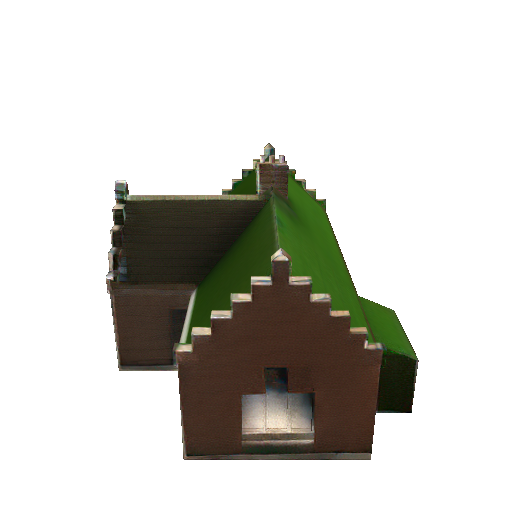} &
        \includegraphics[width=0.15\linewidth]{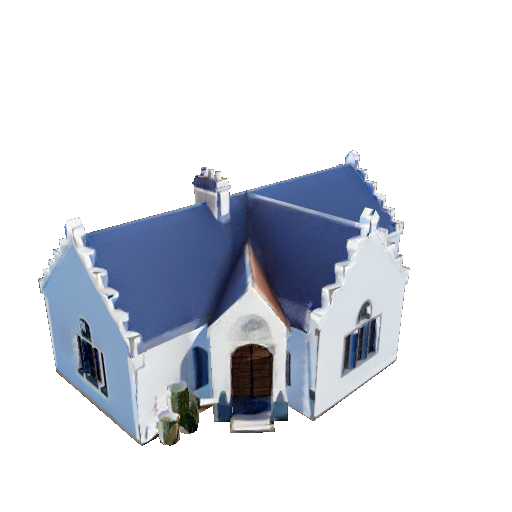} &
        \includegraphics[width=0.15\linewidth]{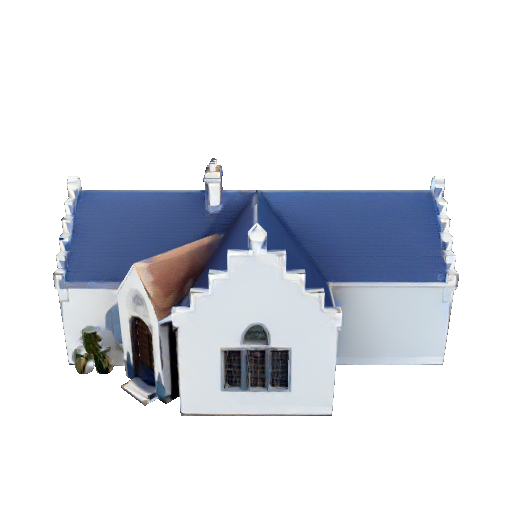} &
        \includegraphics[width=0.15\linewidth]{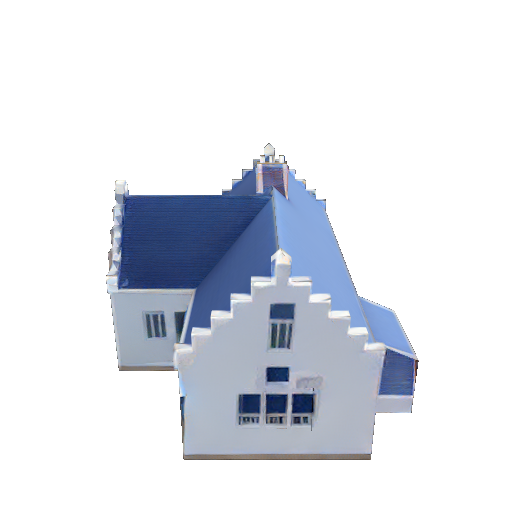} \\
                \multicolumn{3}{c}{``minecraft house, bricks, rock, grass, stone" } & \multicolumn{3}{c}{``colonial style house, white walls, blue ceiling" }\\
        
    \end{tabular}}
    
    \caption{Gallery of meshes textured by \ourmodel. 
    }
    \label{fig:even_more_results}
\end{figure*}

We provide in the supplementary video multi-view renderings of all examples we show in the main paper. Further, in this document, we provide additional results of our method in Fig.~\ref{fig:even_more_results_cnet} and Fig.~\ref{fig:even_more_results}, and comparison to two additional baselines in Fig.~\ref{fig:new_baseline} as described in Sec.~\ref{sec:Additional_baseline}.

\subsection{Additional Baselines} \label{sec:Additional_baseline}
We include two additional methods for qualitative comparison. First is stable-dreamfusion \cite{stable-dreamfusion}, a community-implemented version of Dreamfusion \cite{poole2022dreamfusion} that replaces the proprietary Imagen diffusion model with Stable Diffusion 1.4. Although stable-dreamfusion is a text-to-3D method, not text-to-texture, we include it in our experiments because it is a recently released method and it illustrates the difficulty of jointly synthesizing geometry and texture. We use the default hyperparameters provided in this \href{https://github.com/ashawkey/stable-dreamfusion}{repository}\footnote{https://github.com/ashawkey/stable-dreamfusion}, which performs SDS optimization for $10{,}000$ iterations, with a classifier free guidance weight of 100. The second baseline method is the latent-painter variant of latent-nerf~\cite{metzer2022latent}, for synthesizing textures on an input mesh. Latent-painter performs the same task as us, namely text and geometry-conditioned texture generation, but it does so using the SDS optimization, akin to \cite{poole2022dreamfusion}. We include this method as it was recently the state-of-the-art in texture synthesis with 2D image priors. We use the default hyperparameters provided with this \href{https://github.com/eladrich/latent-nerf}{repository}\footnote{https://github.com/eladrich/latent-nerf}, which performs $5,000$ iterations of SDS optimization, also with a classifier free guidance weight of 100.

\begin{figure}[h]
    \centering
    \includegraphics[width=0.35\linewidth]{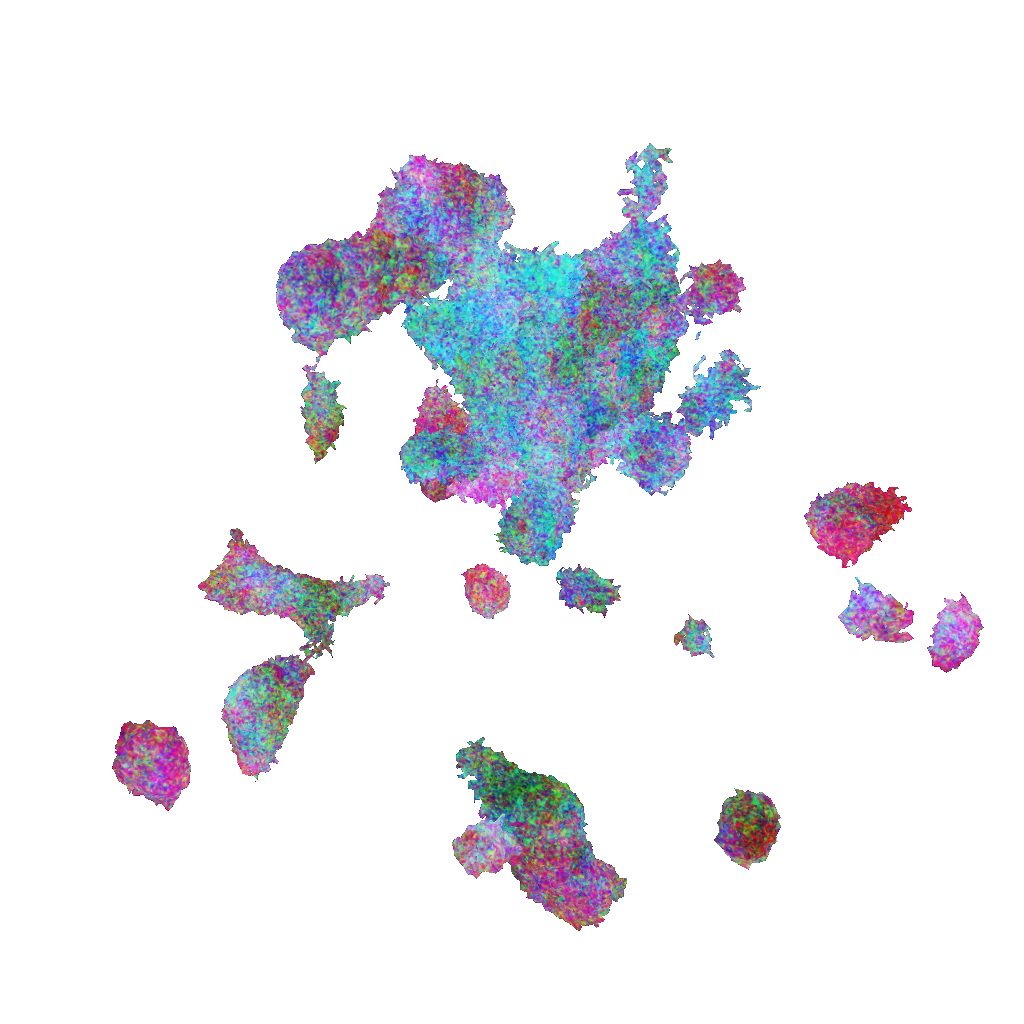}
    \caption{Example result of Stable-DreamFusion where the geometry did not converge properly. Prompt is ``ambulance, white paint with red accents".}
    \label{fig:failed_to_converge}
\end{figure}

Results from these two baselines, along with results from \ourmodel on the same prompts, can be found in Fig.~\ref{fig:new_baseline}. Stable Dreamfusion failed to converge at all for most prompts in our dataset (\eg Fig.~\ref{fig:failed_to_converge}), so we selected prompts where Stable DreamFusion did produce reasonable geometry for visualization. This outcome highlights the fragility of optimizing 3D geometry and texture jointly. We find that Latent-Painter often produced over-saturated colors in the texture due to the use of the SDS optimization with high guidance weights. Furthermore, we find significant artifacts in Latent-Painter results that are reminiscent of incorrect UV mapping. This artifact is in fact due to Latent-Painter applying Stable Diffusion's decoder to the latent texture map directly in texture space, thereby creating artifacts at all boundaries of UV islands. Our method does not suffer from the same issue because we apply the decoder to multiview latent images, making our method agnostic to the underlying UV parameterization. 

\vspace{-3mm}


\subsection{Runtime Comparison}
We compare the runtime of \ourmodel to baselines running on a workstation with a single NVIDIA RTX A6000 GPU in Tab.~\ref{tb:runtime}. We separately measure the runtime of our method under two different camera configurations (see Appendix Section~\ref{sec:camera} for details of the camera configuration). We find \ourmodel to be an order of magnitude faster than methods that rely on optimizing a neural representation with SDS (17.7x w.r.t stable-dreamfusion and 10x w.r.t. Latent Painter). Our runtime is similar to the concurrent work of TEXTure (2.9 min), whose runtime falls between the 9 camera configuration of our method (2.2 min) and 24 camera configuration of our method (6.2 min). Of the 2.2 min duration, 76 seconds are spent on the first round of SIMS, 22 s on the second round, and 34 s on optimizing the neural color field. 

\begin{table}
\small
\centering
\begin{tabular}{l c } 
\toprule
Method & Runtime \\
\midrule
stable-dreamfusion  & 39 min    \\
Latent Painter &  22 min  \\
TEXTure (reported in \cite{richardson2023texture})   & 5 min        \\
TEXTure (ran on our setup) & 2.9 min \\
\midrule
\ourmodel (24 cameras) & 6.2 min \\
\ourmodel (9 cameras) & 2.2 min \\
\bottomrule
\end{tabular}
\caption{Runtime comparison: wall clock time elapsed to synthesize one sample 
}
\vspace{-0.1cm}
\label{tb:runtime}
\end{table}

\begin{figure}[t]
    \centering
    \includegraphics[width=0.8\linewidth]{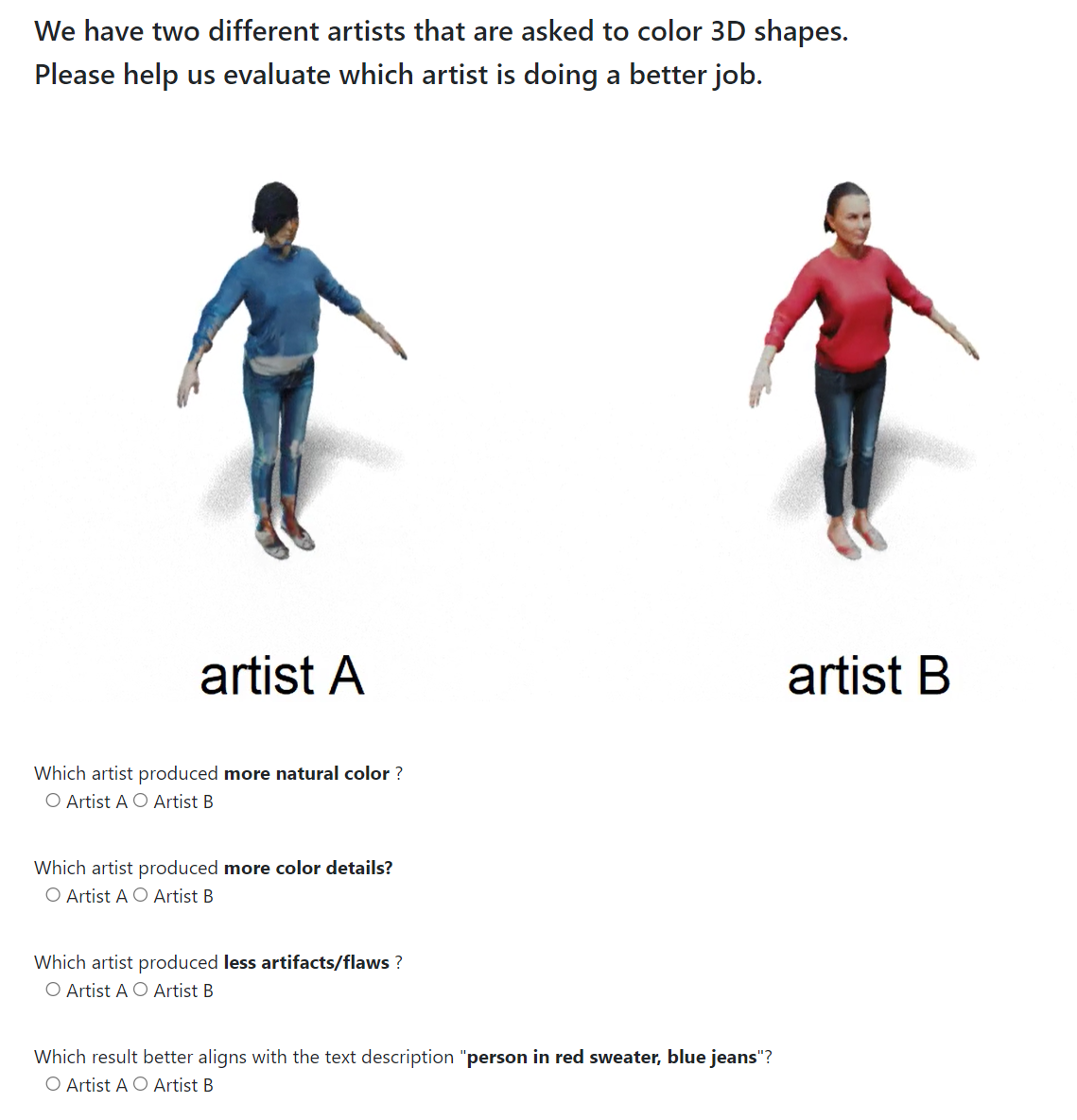}
    \caption{Screenshot of example user study screen}
    \label{fig:user_study_screenshot}
\end{figure}

\section{Experiment details}

\subsection{User study details}
We conduct a user study using Amazon Mechanical Turk \url{https://www.mturk.com/}. We ask each survey participant to look at one pair of texturing results generated by TEXTure and \ourmodel according to the same prompt, displayed side-by-side in random left-right order, and answer four questions.  For each prompt, we show the survey to 3 participants. We then aggregate the results over all responses. A screenshot of one such survey is shown in Fig.~\ref{fig:user_study_screenshot}.

\subsection{Dataset description}
We collect 35 meshes from various sources. A complete list can be found in Tab.~\ref{tb:data1} and Tab.~\ref{tb:data2}. Objects from shapenet are selected from ShapeNetCore.v1, obtained under the \href{https://shapenet.org/terms}{ShapeNet license}\footnote{https://shapenet.org/terms}. One Human model is obtained from Text2Mesh \href{https://github.com/threedle/text2mesh/tree/main/data/source_meshes}{repository}\footnote{https://github.com/threedle/text2mesh/tree/main/data/source\_meshes}. Objects ``house" and ``casa" are obtained for free from Turbosquid with permissive licensing. ``bunny" and ``dragon" are obtained from \href{http://graphics.stanford.edu/data/3Dscanrep/}{Stanford 3D scans}\footnote{http://graphics.stanford.edu/data/3Dscanrep/}. ``Hermanaubis" and ``Provost" are obtained from \href{https://threedscans.com/}{3D scans}\footnote{https://threedscans.com/}, which are shared freely without copyright restrictions. All other objects are obtained under appropriate commercial licenses.

\begin{table*}
\small
\centering
\begin{tabular}{l c c c} 
\toprule
Object & Source & Description & Prompts \\
\midrule
1a64bf1e658652ddb11647ffa4306609 & shapenet & SUV & \begin{tabular}{c}
    ``lamborghini urus" \\
      ``pink porsche cayenne" \\
      ``white mercedes benz SUV" \\
      ``green ambulance with red cross"
\end{tabular}\\ \midrule
1a7b9697be903334b99755e16c4a9d21 & shapenet & coupe & \begin{tabular}{c}
    ``silver porsche 911"\\
      ``blue bmw m5 with white stripes"\\
      ``red ferrari with orange headlights"\\
      ``beautiful yellow sports car"
\end{tabular}\\ \midrule
1a48d03a977a6f0aeda0253452893d75 & shapenet & pickup truck & \begin{tabular}{c}
   ``black pickup truck"\\
      ``old toyota pickup truck"\\
      ``red pickup truck with black trunk"
\end{tabular}\\ \midrule
133c16fc6ca7d77676bb31db0358e9c6 & shapenet & luggage box & \begin{tabular}{c}
        ``blue luggage box"\\
      ``black luggage with a yellow smiley face"
\end{tabular}\\ \midrule
1b9ef45fefefa35ed13f430b2941481 & shapenet & handbag & \begin{tabular}{c}
       ``white handbag"\\
      ``turquoise blue handbag"\\
      ``black handbag with gold trims"
\end{tabular}\\ \midrule
54cd45b275f551b276bb31db0358e9c6 & shapenet & backpack & \begin{tabular}{c}
      ``red backpack"\\
      ``camper bag, camouflage"\\
      ``black backpack with red accents"
\end{tabular}\\ \midrule
e49f6ae8fa76e90a285e5a1f74237618 & shapenet & handbag & \begin{tabular}{c}
     ``crocodile skin handbag"\\
      ``blue handbag with silver trims"\\
      ``linen fabric handbag"
\end{tabular}\\ \midrule
2c6815654a9d4c2aa3f600c356573d21 & shapenet & lounge chair & \begin{tabular}{c}
          ``leather lounge chair"\\
      ``red velvet lounge chair"
\end{tabular}\\ \midrule
2fa970b5c40fbfb95117ae083a7e54ea & shapenet & two-seat sofa & \begin{tabular}{c}
  ``soft pearl fabric sofa"\\
      ``modern building in the shape of a sofa"
\end{tabular}\\ \midrule
5bfee410a492af4f65ba78ad9601cf1b & shapenet & bar stool & \begin{tabular}{c}
      ``yellow plastic stool with white seat"\\
      ``silver metallic stool"
\end{tabular}\\ \midrule
97cd4ed02e022ce7174150bd56e389a8 & shapenet & dinning chair & \begin{tabular}{c}
      ``wooden dinning chair with leather seat"\\
      ``cast iron dinning chair"
\end{tabular}\\ \midrule
5b04b836924fe955dab8f5f5224d1d8a & shapenet & bus & \begin{tabular}{c}
      ``yellow school bus"
\end{tabular}\\ \midrule
7fc729def80e5ef696a0b8543dac6097 & shapenet & taxi sedan & \begin{tabular}{c}
            ``new york taxi, yellow cab"\\
      ``taxi from tokyo, black toyota crown"
\end{tabular}\\ \midrule
85a8ee0ef94161b049d69f6eaea5d368 & shapenet & van & \begin{tabular}{c}
       ``green ambulance with red cross"\\
      ``ambulance, white paint with red accents"\\
      ``pink van with blue top"
\end{tabular}\\ \midrule
a3d77c6b58ea6e75e4b68d3b17c43658 & shapenet & beetle & \begin{tabular}{c}
       ``old and rusty volkswagon beetle"\\
      ``red volkswagon beetle, cartoon style"
\end{tabular}\\ \midrule
b4a86e6b096bb93eb7727d322e44e79b & shapenet & pickup truck & \begin{tabular}{c}
      ``classic red farm truck"\\
      ``farm truck from cars movie, brown, rusty"
\end{tabular}\\ \midrule
fc86bf465674ec8b7c3c6f82a395b347 & shapenet & sports car & \begin{tabular}{c}
     ``batmobile"\\
      ``blue bugatti chiron"
\end{tabular} \\ \midrule
person & \href{https://github.com/threedle/text2mesh}{Text2Mesh} & Human model & \begin{tabular}{c}
    ``white humanoid robot, movie poster, \\ 
    main character of a science fiction movie"\\
      ``comic book superhero, red body suit"\\
      ``white humanoid robot, movie poster, \\
      villain character of a science fiction movie"
\end{tabular} \\
\bottomrule
\end{tabular}
\caption{Description of all geometries used in our dataset, (continued in Tab.~\ref{tb:data2})
}
\vspace{-0.1cm}
\label{tb:data1}
\end{table*}

\begin{table*}
\small
\centering
\resizebox{.97\linewidth}{!}{
\setlength{\tabcolsep}{3pt}
\begin{tabular}{l c c c} 
\toprule
Object & Source & Description & Prompts \\
\midrule

rp\_alvin\_rigged\_003\_MAYA & \href{https://renderpeople.com/3d-people/alvin-rigged-003/}{Renderpeople} & Human model & \begin{tabular}{c}
      ``person wearing black shirt and white pants"\\
      ``person wearing white t-shirt with a peace sign"
\end{tabular} \\ \midrule
rp\_alexandra\_rigged\_004\_MAYA & \href{https://renderpeople.com/3d-people/alexandra-rigged-004/}{Renderpeople} & Human model& \begin{tabular}{c}
      ``person in red sweater, blue jeans"\\
      ``person in white sweater with a red logo, yoga pants"
\end{tabular} \\ \midrule
rp\_adanna\_rigged\_007\_MAYA & \href{https://renderpeople.com/3d-people/adanna_rigged_007/}{Renderpeople} & Human model & \begin{tabular}{c}
    ``nunn in a black dress"\\
      ``nunn in a white dress, black headscarf"
\end{tabular} \\ \midrule
rp\_aaron\_rigged\_001\_MAYA & \href{https://renderpeople.com/3d-people/rp_aaron_rigged_001/}{Renderpeople} & Human model & \begin{tabular}{c}
        ``railroad worker wearing high-vis vest"\\
      ``biker wearing red jacket and black pants"
\end{tabular} \\ \midrule
Age49-LoganWade & \href{https://triplegangers.com/search-products/logan-wade}{Tripleganger} & Human head  & \begin{tabular}{c}
        ``oil painting of a bald, middle aged banker\\
        with pointed moustache"\\
      ``moai stone statue with green moss on top"\\
      ``portrait photo of abraham lincoln, full color"
\end{tabular} \\ \midrule
Age26-AngelicaCollins & \href{https://triplegangers.com/search-products/angelica-collins}{Tripleganger} & Human head  & \begin{tabular}{c}
 ``Portrait of a humanoid robot, futuristic, science fiction"\\
      ``blonde girl with green eyes, hair in tied a bun, \\ 
      anime illustration, portrait"\\
      ``blonde girl with green eyes, hair in tied a bun, \\ 
      DSLR portrait photo"
\end{tabular} \\ \midrule
house & \href{https://www.turbosquid.com/3d-models/3d-lowpoly-cartoon-medieval-house-polygonal-2028066}{Turbosquid} & Medieval house  & \begin{tabular}{c}
    ``medieval celtic House, stone bricks, wooden roof"\\
      ``minecraft house, bricks, rock, grass, stone"\\
      ``colonial style house, white walls, blue ceiling"
\end{tabular} \\ \midrule
casa & \href{https://www.turbosquid.com/3d-models/house-3d-model-1412380}{Turbosquid} & house in the sea & \begin{tabular}{c}
    ``white house by the dock, green ceiling, cartoon style" \\
      ``minecraft house, bricks, rock, grass, stone" \\
      ``white house by the dock, green ceiling, impressionist painting"
\end{tabular} \\ \midrule

1073771 & \href{https://www.turbosquid.com/}{Turbosquid} & rabbit & \begin{tabular}{c}
    ``brown rabbit"\\
      ``purple rabbit"\\
      ``tiger with yellow and black stripes"
\end{tabular} \\ \midrule

1106184 & \href{https://www.turbosquid.com/}{Turbosquid} & cartoon dog & \begin{tabular}{c}
``cartoon dog"\\
      ``lion dance, red and green"\\
      ``brown bull dog"
\end{tabular} \\ \midrule

1117733 & \href{https://www.turbosquid.com/}{Turbosquid} & goat & \begin{tabular}{c}
 ``brown mountain goat"\\
      ``black goat with white hoofs"\\
      ``milk cow"
\end{tabular} \\ \midrule

1281334 & \href{https://www.turbosquid.com/}{Turbosquid} & cartoon cow & \begin{tabular}{c}
``cartoon milk cow"\\
      ``giant panda"
\end{tabular} \\ \midrule

1367642 & \href{https://www.turbosquid.com/}{Turbosquid} & cartoon fox & \begin{tabular}{c}
  ``cartoon fox"\\
      ``brown wienner dog"\\
      ``white fox"
\end{tabular} \\ \midrule

bunny & \href{http://graphics.stanford.edu/data/3Dscanrep/}{Stanford 3D Scans} & bunny & \begin{tabular}{c}
    ``white bunny" 
\end{tabular} \\ \midrule
dragon & \href{http://graphics.stanford.edu/data/3Dscanrep/}{Stanford 3D Scans} & dragon & \begin{tabular}{c}
          ``black and white dragon in chinese ink art style"\\
      ``cartoon dragon, red and green"
\end{tabular} \\ \midrule
Hermanubis & \href{https://threedscans.com/}{3D scans} & statue & \begin{tabular}{c}
         ``sandstone statue of hermanubis" \\
      ``portrait of greek-egyptian deity hermanubis, lapis skin and gold clothing"
\end{tabular} \\ \midrule
Provost & \href{https://threedscans.com/}{3D scans} & statue & \begin{tabular}{c}
        ``portrait of Provost, oil paint"\\
      ``marble statue of Provost"
\end{tabular} \\ \midrule
\bottomrule
\end{tabular}
}
\caption{Description of all geometries used in our dataset continued.
}
\vspace{-0.1cm}
\label{tb:data2}
\end{table*}

\end{document}